\documentclass{article}

\usepackage[final]{neurips_2018}

\usepackage[utf8]{inputenc} 
\usepackage[T1]{fontenc}    
\usepackage[pagebackref=false,breaklinks=true,colorlinks,bookmarks=false,citecolor=blue]{hyperref}

\usepackage{url}            
\usepackage{booktabs}       
\usepackage{amsfonts}       
\usepackage{nicefrac}       
\usepackage{microtype}      

\usepackage{amsmath}
\usepackage{amsthm}
\usepackage{amssymb}
\usepackage{bbm}
\usepackage{paralist}
\usepackage{enumitem}
\usepackage{algorithm}
\usepackage{algorithmic}
\usepackage{xcolor}
\usepackage{mathtools}

\usepackage{subcaption}

\newtheorem{theorem}{Theorem}

\newtheorem{lemma}{Lemma}

\newcommand{\ep}{\hfill $\Box$}

\renewcommand{\cite}{\citep}
\DeclareMathOperator{\EXP}{\mathbb{E}}
\renewcommand{\Pr}{\mathbb{P}}
\renewcommand{\star}{*}
\newcommand{\set}[1]{\mathcal{#1}}

\newcommand{\kl}{\textnormal{KL}}

\newcommand{\TN}{\textnormal}

\DeclareMathOperator*{\argmin}{arg\,min}
\DeclareMathOperator*{\argmax}{arg\,max} 

\newcommand{\est}{\textnormal{est}}
\newcommand{\mnt}{\textnormal{mnt}}
\newcommand{\xpt}{\textnormal{xpt}}
\newcommand{\xpr}{\textnormal{xpr}}

\title{Exploration in Structured Reinforcement Learning}

\author{
Jungseul Ok\\
KTH, EECS  \\
Stockholm, Sweden\\
\texttt{ockjs@illinois.edu}
\And
 Alexandre Proutiere  \\
  KTH, EECS \\
Stockholm, Sweden\\
\texttt{alepro@kth.se}
 \And
   Damianos Tranos \\
  KTH, EECS \\
  Stockholm, Sweden\\
\texttt{tranos@kth.se}
}

\begin{document}

\maketitle


\begin{abstract}
We address reinforcement learning problems with finite state and action spaces where the underlying MDP has some known structure that could be potentially exploited to minimize the exploration rates of suboptimal (state, action) pairs. For any arbitrary structure, we derive problem-specific regret lower bounds satisfied by any learning algorithm. These lower bounds are made explicit for unstructured MDPs and for those whose transition probabilities and average reward functions are Lipschitz continuous w.r.t. the state and action. For Lipschitz MDPs, the bounds are shown not to scale with the sizes $S$ and $A$ of the state and action spaces, i.e., they are smaller than $c\log T$ where $T$ is the time horizon and the constant $c$ only depends on the Lipschitz structure, the span of the bias function, and the minimal action sub-optimality gap. This contrasts with unstructured MDPs where the regret lower bound typically scales as $SA\log T$. We devise DEL (Directed Exploration Learning), an algorithm that matches our regret lower bounds. We further simplify the algorithm for Lipschitz MDPs, and show that the simplified version is still able to efficiently exploit the structure.

\end{abstract}


%


\section{Introduction}

Real-world Reinforcement Learning (RL) problems often concern dynamical systems with {\it large} state and action spaces, which make the design of efficient algorithms extremely challenging. This difficulty is well illustrated by the known {\it regret} fundamental limits. The regret compares the accumulated reward of an optimal policy (aware of the system dynamics and reward function) to that of the algorithm considered, and it quantifies the loss incurred by the need of exploring sub-optimal (state, action) pairs to learn the system dynamics and rewards. In online RL problems with undiscounted reward, regret lower bounds typically scale as $SA\log T$ or $\sqrt{SAT}$\footnote{The first lower bound is asymptotic in $T$ and problem-specific, the second is minimax. We ignore here for simplicity the dependence of these bounds in the diameter, bias span, and action sub-optimality gap.}, where $S$, $A$, and $T$ denote the sizes of the state and action spaces and the time horizon, respectively. Hence, with large state and action spaces, it is essential to identify and exploit any possible structure existing in the system dynamics and reward function so as to minimize exploration phases and in turn reduce regret to reasonable values. Modern RL algorithms actually implicitly impose some structural properties either in the model parameters (transition probabilities and reward function, see e.g. \cite{ortner2012online}) or directly in the $Q$-function (for discounted RL problems, see e.g. \cite{mnih015}. Despite the successes of these recent algorithms, our understanding of structured RL problems remains limited. 

In this paper, we explore structured RL problems with finite state and action spaces. We first derive problem-specific regret lower bounds satisfied by any algorithm for RL problems with any arbitrary structure. These lower bounds are instrumental to devise algorithms optimally balancing exploration and exploitation, i.e., achieving the regret fundamental limits. A similar approach has been recently applied with success to stochastic bandit problems, where the average reward of arms exhibits structural properties, e.g. unimodality \cite{Combes2014}, Lipschitz continuity \cite{magureanu2014}, or more general properties \cite{combesnips17}. Extending these results to RL problems is highly non trivial, and to our knowledge, this paper is the first to provide problem-specific regret lower bounds for structured RL problems. Although the results presented here concern ergodic RL problems with undiscounted reward, they could be easily generalized to discounted problems (under an appropriate definition of regret).   

Our contributions are as follows:

1. For ergodic structured RL problems, we derive problem-specific regret lower bounds. The latter are valid for any structure (but are structure-specific), and for unknown system dynamics and reward function. \\
2. We analyze the lower bounds for unstructured MDPs, and show that they scale at most as ${(H+1)^2\over \delta_{\min}}SA\log T$, where $H$ and $\delta_{\min}$ represent the span of the bias function and the minimal state-action sub-optimality gap, respectively. These results extend previously known regret lower bounds derived in the seminal paper \cite{burnetas1997optimal} to the case where the reward function is unknown. \\
3. We further study the regret lower bounds in the case of Lipschitz MDPs. Interestingly, these bounds are shown to scale at most as $\frac{(H+1)^3}{\delta^2_{\min}} S_{\textnormal{lip}} A_{\textnormal{lip}}\log T$ where $S_{\textnormal{lip}}$ and $A_{\textnormal{lip}}$ only depend on the Lipschitz properties of the transition probabilities and reward function. This indicates that when $H$ and $\delta_{\min}$ do not scale with the sizes of the state and action spaces, we can hope for a regret growing logarithmically with the time horizon, and independent of $S$ and $A$. \\
4. We propose DEL, an algorithm that achieves our regret fundamental limits for any structured MDP. DEL is rather complex to implement since it requires in each round to solve an optimization problem similar to that providing the regret lower bounds. Fortunately, we were able to devise simplified versions of DEL, with regret scaling at most as ${(H+1)^2\over \delta_{\min}}SA\log T$ and $\frac{(H+1)^3}{\delta^2_{\min}} S_{\textnormal{lip}} A_{\textnormal{lip}}\log T$ for unstructured and Lipschitz MDPs, respectively. In absence of structure, DEL, in its simplified version, does not require to compute action indexes as done in OLP \cite{bartlett08}, and yet achieves similar regret guarantees without the knowledge of the reward function. DEL, simplified for Lipschitz MDPs, only needs, in each step, to compute the optimal policy of the estimated MDP, as well as to solve a simple linear program.  \\
5. Preliminary numerical experiments (presented in the appendix) illustrate our theoretical findings. In particular, we provide examples of Lipschitz MDPs, for which the regret under DEL does not seem to scale with $S$ and $A$, and significantly outperforms algorithms that do not exploit the structure.

\section{Related Work}

Regret lower bounds have been extensively investigated for unstructured ergodic RL problems. \cite{burnetas1997optimal} provided a problem-specific lower bound similar to ours, but only valid when the reward function is known. Minimax regret lower bounds have been studied e.g. in \cite{auer2009near} and \cite{regal}: in the worst case, the regret has to scale as $\sqrt{DSAT}$ where $D$ is the diameter of the MDP. In spite of these results, regret lower bounds for unstructured RL problems are still attracting some attention, see e.g. \cite{osband2016lower} for insightful discussions. To our knowledge, this paper constitutes the first attempt to derive regret lower bounds in the case of structured RL problems. Our bounds are asymptotic in the time horizon $T$, but we hope to extend them to finite time horizons using similar techniques as those recently used to provide such bounds for bandit problems \cite{garivier193}. These techniques address problem-specific and minimax lower bounds in a unified manner, and can be leveraged to derive minimax lower bounds for structured RL problems. However we do not expect minimax lower bounds to be very informative about the regret gains that one may achieve by exploiting a structure (indeed, the MDPs leading to worst-case regret in unstructured RL comply to many structures).

There have been a plethora of algorithms developed for ergodic unstructured RL problems. We may classify these algorithms depending on their regret guarantees, either scaling as $\log T$ or $\sqrt{T}$. In absence of structure, \cite{burnetas1997optimal} developed an asymptotically optimal, but involved, algorithm. This algorithm has been simplified in \cite{bartlett08}, but remains more complex than our proposed algorithm. Some algorithms have finite-time regret guarantees scaling as $\log T$ \cite{ucrl}, \cite{auer2009near}, \cite{filippi}. For example, the authors of \cite{filippi} propose KL-UCRL an extension of UCRL \cite{ucrl} with regret bounded by ${D^2S^2A\over\delta_{\min}}\log T$. Having finite-time regret guarantees is arguably desirable, but so far this comes at the expense of a much larger constant in front of $\log T$. Algorithms with regret scaling as $\sqrt{T}$ include UCRL2 \cite{auer2009near}, KL-UCRL with regret guarantees $\tilde{O}(DS\sqrt{AT})$, REGAL.C \cite{regal} with guarantees $\tilde{O}(HS\sqrt{AT})$. Recently, the authors of \cite{agrawal2017} managed to achieve a regret guarantee of $\tilde{O}(D\sqrt{SAT})$, but only valid when $T\ge S^5A$. 

Algorithms devised to exploit some known structure are most often applicable to RL problems with continuous state or action spaces. Typically, the transition probabilities and reward function are assumed to be smooth in the state and action, typically Lipschitz continuous \cite{ortner2012online}, \cite{lakshmanan2015improved}. The regret then needs to scale as a power of $T$, e.g. $T^{2/3}$ in \cite{lakshmanan2015improved} for 1-dimensional state spaces. An original approach to RL problems for which the transition probabilities belong to some known class of functions was proposed in \cite{eluder}. The regret upper bounds derived there depend on the so-called Kolmogorov and eluder dimensions, which in turn depend on the chosen class of functions. Our approach to design learning algorithms exploiting the structure is different from all aforementioned methods, as we aim at matching the problem-specific minimal exploration rates of sub-optimal (state, action) pairs.  
  

\section{Models and Objectives}

We consider an MDP $\phi=(p_\phi,q_\phi)$ with finite state and action spaces $\set{S}$ and $\set{A}$ of respective cardinalities $S$ and $A$. $p_\phi$ and $q_\phi$ are the transition and reward kernels of $\phi$. Specifically, when in state $x$, taking action $a$, the system moves to state $y$ with probability $p_\phi(y|x,a)$, and a reward drawn from distribution $q_\phi(\cdot | x,a)$ of average $r_\phi(x,a)$ is collected. The rewards are bounded, w.l.o.g., in $[0,1]$. We assume that for any $(x,a)$, $q_\phi(\cdot |x,a)$ is absolutely continuous w.r.t. some measure $\lambda$ on $[0,1]$\footnote{$\lambda$ can be the Lebesgue measure; alternatively, if rewards take values in $\{0,1 \}$, $\lambda$ can be the sum of Dirac measures at 0 and 1.}. 

The random vector $Z_t:=(X_t, A_t, R_t)$ represents the state, the action, and the collected reward at step $t$. A policy $\pi$ selects an action, denoted by $\pi_t(x)$, in step $t$ when the system is in state $x$ based on the history captured through $\set{H}_t^\pi$, the $\sigma$-algebra generated by $(Z_1,\ldots, Z_{t-1},X_t)$ observed under $\pi$: $\pi_t(x)$ is $\set{H}_t^\pi$-measurable. We denote by $\Pi$ the set of all such policies. 

{\bf Structured MDPs.} The MDP $\phi$ is initially unknown. However we assume that $\phi$
belongs to some well specified set $\Phi$ which may encode a known structure of the MDP.
The knowledge of $\Phi$ can be exploited to devise (more) efficient policies. The results derived in this paper are valid under {\it any} structure, but we give a particular attention to the cases of
\\
 {\it (i) Unstructured MDPs}: $\phi \in \Phi$ if for all $(x,a)$, $p_\phi (\cdot \mid x,a) \in \set{P}(\set{S})$ and
  $q_\phi (\cdot \mid x,a) \in \set{P}([0,1])$\footnote{$\set{P}(\set{S})$ is the set of distributions on $\set{S}$ and $\set{P}([0,1])$ is the set of distributions on $[0,1]$, absolutely continuous w.r.t. $\lambda$.}; \\
 {\it (ii) Lipschitz MDPs}: $\phi \in \Phi$ if $p_\phi(\cdot |x,a)$ and $r_\phi(x,a)$ 
are Lipschitz-continuous w.r.t. $x$ and $a$ in some metric space (we provide a precise definition in the next section).

{\bf The learning problem.} The expected cumulative reward up to step $T$ of a policy $\pi \in \Pi$ when the system starts in state $x$ is $V_T^\pi(x) := \EXP_x^\pi[\sum_{t=1}^T R_t],$ where $\EXP_x^\pi [\cdot ]$ denotes the expectation under policy $\pi$ given that $X_1=x$. Now assume that the system starts in state $x$ and evolves according to the initially unknown MDP $\phi\in \Phi$ for given structure $\Phi$, the objective is to devise a policy $\pi\in \Pi$ maximizing  $V_T^\pi(x)$ or equivalently, minimizing the regret $R_T^\pi(x)$ up to step $T$ defined as the difference between the cumulative reward of an optimal policy and that obtained under $\pi$:
\begin{align*}
R_T^\pi(x) := V^\star_T(x) - V^\pi_T(x)
\end{align*}
where $V^\star_T(x) := \sup_{ \pi \in \Pi} V^\pi_T (x)$.

{\bf Preliminaries and notations.} 
Let $\Pi_{D}$ be the set of stationary (deterministic) policies, i.e. when in state $X_t=x$, $f\in \Pi_{D}$ selects an action $f(x)$ independent of $t$. 
$\phi$ is communicating if each pair of states are connected by some policy.
Further, $\phi$ is ergodic if under any stationary policy, the resulting Markov chain $(X_t)_{t\ge 1}$ is irreducible.
For any communicating $\phi$ and any policy $\pi\in \Pi_D$, we denote by $g^\pi_\phi(x)$ the {\it gain} of $\pi$ (or long-term average reward) started from initial state $x$: $g_\phi^\pi (x):=\lim_{T\to\infty}{1\over T}V_T^\pi(x)$. 
 We denote by $\Pi^\star(\phi)$ the set of stationary policies with maximal gain: $\Pi^\star(\phi):=\{ f \in \Pi_{D}: g_\phi^f (x)= g_\phi^\star(x)~\forall x \in \set{S}\}$, where $g_\phi^\star (x):=\max_{\pi \in \Pi} g_\phi^\pi(x)$. If $\phi$ is communicating, the maximal gain is constant and denoted by $g^*_\phi$.
 The {\it bias function} $h_\phi^f$ of $f\in \Pi_D$ is defined by $h_\phi^f(x):=\TN{{\em C}-}\lim_{T\to\infty} \EXP_x^f[\sum_{t = 1}^\infty (R_t  - g_\phi^f(X_t))]$, and quantifies the advantage of starting in state~$x$. We denote by $\mathbf{B}^a_\phi$ and $\mathbf{B}^*_\phi$, respectively, the Bellman operator under action $a$ and the optimal Bellman operator under $\phi$. They are defined by: for any $h: \set{S} \mapsto \mathbb{R}$ and $x \in \set{S}$,
\begin{align*}
(\mathbf{B}^a_\phi h)(x) :=  r_\phi(x,a) + \sum_{y \in \set{S}} p_\phi(y| x,a) h(y) \quad \text{and} \quad
(\mathbf{B}^*_\phi h)(x) := \max_{a \in \set{A}} (\mathbf{B}^a_\phi h)(x) \;.
\end{align*}
Then for any $f \in \Pi_{D}$, $g_\phi^f$ and $h_\phi^f$ satisfy the {\it evaluation equation}: for all state $x\in\set{S}$, $g_\phi^f(x)+ h_\phi^f(x) = (\mathbf{B}^{f(x)}_\phi  h_\phi^f) (x)$. Furthermore, $f \in \Pi^\star(\phi)$ if and only if 
$g_\phi^f$ and $h_\phi^f$ verify the {\it optimality equation}: 
\begin{align*}
g_\phi^f(x) + h_\phi^f(x) = (\mathbf{B}^{*}_\phi  h_\phi^{f}) (x) \;.
\end{align*}
We denote by $h_\phi^\star$ the bias function of an optimal stationary policy\footnote{In case of $h^*_\phi$ is not unique,
we arbitrarily select an optimal stationary policy and define $h^*_\phi$.}, and by $H$ its span $H:=\max_{x,y}h_\phi^\star(x)-h_\phi^\star(y)$. For $x\in \set{S}$, $h: \set{S} \mapsto \mathbb{R}$, and $\phi\in\Phi$, let $\set{O}(x;h, \phi) = \{a \in \set{A}: (\mathbf{B}^\star_\phi h)(x) =(\mathbf{B}^a_\phi h)(x) \}$. 
For ergodic $\phi$, $h^*_\phi$ is unique up to an additive constant.
Hence, for ergodic $\phi$, the set of optimal actions in state $x$ under $\phi$ is $\set{O}(x; \phi) := \set{O}(x; h^\star_\phi, \phi)$, and $\Pi^*(\phi) = \{f \in \Pi_D : f(x) \in \set{O}(x;\phi) ~\forall x \in \set{S}\}$. Finally, we define for any state $x$ and action~$a$,
\begin{align*}
\delta^*(x,a;\phi) :=  (\mathbf{B}_\phi^\star h_\phi^\star)(x) - 
(\mathbf{B}_\phi^a h_\phi^\star)(x) \;.
\end{align*}
This can be interpreted as the long-term regret obtained by initially selecting action $a$ in state $x$ (and then applying an optimal stationary policy) rather than following an optimal policy. The minimum gap is defined as $\delta_{\min}:= \min_{(x,a): \delta^*(x,a;\phi)>0}\delta^*(x,a;\phi)$.

We denote by $\bar{\mathbb{R}}_+=\mathbb{R}_+\cup \{\infty\}$. The set of MDPs is equipped with the following $\ell_\infty$-norm: $\| \phi -\psi \| := \max_{(x,a) \in \set{S} \times \set{A}} \| \phi(x,a) -\psi(x,a) \|$ where $\| \phi(x,a) -\psi(x,a) \| := |r_\phi (x,a) - r_\psi (x,a)|+\max_{y\in \set{S}} |p_\phi (y \mid x,a) - p_\psi (y \mid x,a)| $.

The proofs of all results are presented in the appendix.



\section{Regret Lower Bounds}

In this section, we present an (asymptotic) regret lower bound satisfied by any {\it uniformly good} learning algorithm. An algorithm $\pi\in \Pi$ is uniformly good if for all ergodic $\phi\in \Phi$, any initial state $x$ and any constant $\alpha>0$, the regret of $\pi$ satisfies $R_T^\pi(x)=o(T^\alpha)$ . 

To state our lower bound, we introduce the following notations.
For $\phi$ and $\psi$, we denote $\phi \ll \psi$ if the kernel of $\phi$ is absolutely continuous w.r.t. that of $\psi$, i.e.,
$\forall \set{E}$, $\Pr_\phi[\set{E}] = 0$ if $\Pr_\psi[\set{E}] = 0$.
For $\phi$ and $\psi$ such that $\phi \ll \psi$ and $(x,a)$, we define the KL-divergence between $\phi$ and $\psi$ in state-action pair $(x,a)$ $\kl_{\phi \mid \psi}(x,a) $ as the KL-divergence between 
the distributions of the next state and collected reward if the state is $x$ and $a$ is selected under these two MDPs:
\begin{align*}
\kl_{\phi \mid \psi}(x,a) 
= \sum_{y\in \set{S}} p_\phi(y | x,a)  \log \frac{p_\phi(y | x,a) }{p_\psi(y | x,a) }  
+
\int_{0}^1 q_\phi(r | x,a) \log \frac{q_\phi(r | x,a)}{q_\psi(r | x,a) }\lambda(dr)  \;.
\end{align*}
We further define the set of {\it confusing} MDPs as:
\begin{align*}
\Delta_{\Phi} (\phi) = \{\psi \in \Phi :  \phi \ll \psi,~ (i) \ \kl_{\phi \mid \psi}(x,a) = 0~  \forall x, \forall a \in \set{O}(x ;\phi);~(ii)\  \Pi^*(\phi) \cap \Pi^*(\psi) = \emptyset \} \;.
\end{align*}
This set consists of MDP $\psi$'s that $(i)$ coincide with $\phi$ for state-action pairs where the actions are optimal (the kernels of $\phi$ and $\psi$ cannot be statistically distinguished under an optimal policy); and such that $(ii)$ the optimal policies under $\psi$ are not optimal under $\phi$. 

\begin{theorem} \label{thm:g-r-lower}
Let $\phi \in \Phi$ be ergodic. For any uniformly good algorithm $\pi \in \Pi$ and for any $x \in \set{S}$, 
\begin{align} \label{eq:g-r-lower}
\liminf_{T \to \infty} \frac{R^\pi_T (x)}{\log T} \ge K_{\Phi}(\phi), 
\end{align}
where $K_{\Phi}(\phi)$ is the value of the following optimization problem:
\begin{align}
& \underset{\eta \in \set{F}_\Phi(\phi) }{\textnormal{inf}} 
~~\sum_{(x,a) \in \set{S}\times \set{A}} \eta(x, a)  \delta^*(x, a;\phi), \label{eq:g-r-obj}
\end{align}
where $\set{F}_\Phi(\phi):=\{\eta \in \bar{\mathbb{R}}_+^{S\times A}: \sum_{(x,a) \in \set{S} \times \set{A}}\eta(x, a) \kl_{\phi\mid \psi}(x,a) \ge 1, \forall \psi \in \Delta_{\Phi} (\phi)\}$.
\end{theorem}

The above theorem can be interpreted as follows. When selecting a sub-optimal action $a$ in state $x$, one has to pay a regret of $\delta^*(x, a;\phi)$. Then the minimal number of times any sub-optimal action $a$ in state $x$ has to be explored scales as $\eta^\star(x,a)\log T$ where $\eta^\star(x,a)$ solves the optimization problem (\ref{eq:g-r-obj}). It is worth mentioning that our lower bound is tight, as we present in Section \ref{sec:algo} an algorithm achieving this fundamental limit of regret. 

The regret lower bound stated in Theorem \ref{thm:g-r-lower} extends the problem-specific regret lower bound derived in \cite{burnetas1997optimal} for unstructured ergodic MDPs with known reward function. Our lower bound is valid for unknown reward function, but also applies to any structure $\Phi$. Note however that at this point, it is only implicitly defined through the solution of \eqref{eq:g-r-obj}, which seems difficult to solve. The optimization problem can actually be simplified, as shown later in this section, by providing useful structural properties of the feasibility set $\set{F}_{\Phi}(\phi)$ depending on the structure considered. The simplification will be instrumental to quantify the gain that can be achieved when optimally exploiting the structure, as well as to design efficient algorithms.  

In the following, the optimization problem: $\inf_{\eta\in {\cal F}} \sum_{(x,a) \in \set{S}\times \set{A}} \eta(x, a)  \delta^*(x, a;\phi)$ is referred to as $P(\phi,\set{F})$; so that $P(\phi,\set{F}_{\Phi}(\phi))$ corresponds to (\ref{eq:g-r-obj}).

The proof of Theorem \ref{thm:g-r-lower} combines a characterization of the regret as a function of the number of times $N_T(x,a)$ up to step $T$ (state, action) pair $(x,a)$ is visited, and of the $\delta^*(x, a;\phi)$'s, and change-of-measure arguments as those recently used to prove in a very direct manner regret lower bounds in bandit optimization problems \cite{kaufmann2016complexity}. More precisely, for any uniformly good algorithm $\pi$, and for any confusing MDP $\psi\in  \Delta_{\Phi} (\phi)$, we show that the exploration rates required to statistically distinguish $\psi$ from $\phi$ satisfy $\liminf_{T\to\infty}{1\over \log T}\sum_{(x,a) \in \set{S} \times \set{A}}\mathbb{E}_{x_1}^\pi[N_T(x, a)] \kl_{\phi\mid \psi}(x,a)\ge 1$ where the expectation is taken w.r.t. $\phi$ given any initial state $x_1$. The theorem is then obtained by considering (hence optimizing the lower bound) all possible confusing MDPs.

\subsection{Decoupled exploration in unstructured MDPs}

In the absence of structure, $\Phi=\{\psi: p_\psi(\cdot |x,a)\in \set{P}(\set{S}), q_\psi(\cdot |x,a)\in  \set{P}([0,1]), \forall (x,a)\}$, and we have:

\begin{theorem} \label{thm:unrelated}
Consider the unstructured model $\Phi$, and let $\phi\in \Phi$ be ergodic. We have:
\begin{align*}
\set{F}_\Phi(\phi)
=
\left\{
\eta \in \bar{\mathbb{R}}_+^{S\times A}:  \forall (x,a)~\text{s.t.}~a \notin \set{O}(x; \phi), \eta(x, a) \kl_{\phi\mid \psi}(x,a) \ge 1, ~
\forall \psi \in \Delta_{\Phi}(x, a; \phi)
\right\} 
\end{align*}
where
$
\Delta_{\Phi}(x,a; \phi) :=
\{\psi \in \Phi  : \kl_{\phi \mid \psi} (y, b) = 0 ~\forall (y,b) \neq (x,a)~\text{and}~ (\mathbf{B}^a_\psi h^*_\phi)(x) > g^*_\phi + h^*_\phi(x)\}.
$
\end{theorem}

The theorem states that in the constraints of the optimization problem (\ref{eq:g-r-obj}), we can restrict our attention to confusing MDPs $\psi$ that are different than the original MDP $\phi$ only for a single state-action pair $(x,a)$. Further note that the condition $(\mathbf{B}^a_\psi h^*_\phi)(x) > g^*_\phi + h^*_\phi(x)$ is equivalent to saying that action $a$ becomes optimal in state $x$ under $\psi$ (see Lemma 1(i) in \cite{burnetas1997optimal}). Hence to obtain the lower bound in unstructured MDPs, we may just consider confusing MDPs $\psi$ which make an initially sub-optimal action $a$ in state $x$ optimal by locally changing the kernels and rewards of $\phi$ at $(x,a)$ only. Importantly, this observation implies that an optimal algorithm $\pi$ must satisfy $\mathbb{E}_{x_1}^\pi[N_T(x,a)] \sim \log T / \inf_{\psi\in \Delta_{\Phi}(x,a; \phi)} \kl_{\phi\mid \psi}(x,a)$. In other words, the required level of exploration of the various sub-optimal state-action pairs are {\it decoupled}, which significantly simplifies the design of optimal algorithms.

To get an idea on how the regret lower bound scales as the sizes of both state and action spaces, we can further provide an upper bound of the regret lower bound. One may easily observe that $\set{F}_{\TN{un}}(\phi) \subset \set{F}_\Phi(\phi)$ where
\begin{align*}
\set{F}_{\TN{un}}(\phi)
=
\left\{
\eta  \in \bar{\mathbb{R}}_+^{S\times A}: \eta(x, a) \left(\frac{\delta^*(x, a;\phi)}{ H+1}\right)^2 \ge 2, ~ 
\forall (x,a)~\text{s.t.}~a \notin \set{O}(x; \phi)
\right\} \;.
\end{align*}
From this result, an upper bound of the regret lower bound is $K_{\textnormal{un}}(\phi):=2{(H+1)^2\over \delta_{\min}}SA\log T$, and we can devise algorithms achieving this regret scaling (see Section \ref{sec:algo}).   

Theorem \ref{thm:unrelated} relies on the following decoupling lemma, actually valid under any structure $\Phi$.

 \begin{lemma} \label{lem:gen-decom} Let $\set{U}_1,\set{U}_2$ be two non-overlapping subsets of the (state, action) pairs such that for all $(x,a)\in \set{U}_0:=\set{U}_1\cup\set{U}_2$, $a\notin \set{O}(x;\phi)$. Define the following three MDPs in $\Phi$ obtained starting from $\phi$ and changing the kernels for (state, action) pairs in $\set{U}_1\cup\set{U}_2$. Specifically, let $(p,q)$ be some transition and reward kernels. For all $(x,a)$, define $\psi_j$, $j\in\{0,1,2\}$ as 
\begin{align*}
(p_{\psi_j}(\cdot | x, a), q_{\psi_j}(\cdot |x,a))  &=
\begin{cases}
(p(\cdot | x, a),q(\cdot |x,a)) & \text{if $(x, a) \in  \set{U}_j$}, \\
(p_{\phi}(\cdot | x, a),q_\phi(\cdot |x,a))  & \text{otherwise}. 
\end{cases} 
\end{align*} 
Then, if $\Pi^*(\phi) \cap \Pi^*(\psi_0) = \emptyset$, then 
$\Pi^*(\phi) \cap \Pi^*(\psi_1) = \emptyset$ or
$\Pi^*(\phi) \cap \Pi^*(\psi_2) = \emptyset$.
 \end{lemma}

\subsection{Lipschitz structure}

Lipschitz structures have been widely studied in the bandit and reinforcement learning literature. We find it convenient to use the following structure, although one could imagine other variants in more general metric spaces. We assume that the state (resp. action) space can be embedded in the $d$ (resp. $d'$) dimensional Euclidian space: $\set{S}\subset [0,D]^d$ and $\set{A}\subset [0,D']^{d'}$. We consider MDPs whose transition kernels and average rewards are Lipschitz w.r.t. the states and actions. Specifically, let $L, L'>0$, $\alpha, \alpha'>0$, and 
\begin{align*}
\Phi=\{\psi: p_\psi(\cdot |x,a)\in \set{P}(\set{S}), q_\psi(\cdot |x,a)\in  \set{P}([0,1]):\text{(L1)-(L2)} \hbox{ hold}, \forall (x,a)\}, 
\end{align*}
where 
\begin{align*}
(L1) & \quad\quad \| p_\psi(\cdot |x,a) - p_\psi(\cdot | x', a') \|_1 
\le L d(x, x')^\alpha + L' d(a,a')^{\alpha'} \;, \\
(L2) & \quad\quad |r_\psi(x,a) - r_\psi(x', a') |  \le L d(x, x')^\alpha + L' d(a,a')^{\alpha'} \;.
\end{align*}
Here $d(\cdot,\cdot)$ is the Euclidean distance, and 
for two distributions $p_1$ and $p_2$ on $\set{S}$ 
we denote by $\| p_1-p_2\|_1\:=\sum_{y\in\set{S}} |p_1(y)-p_2(y)|$. 

\begin{theorem} \label{thm:lip}
For the model $\Phi$ with Lipschitz structure (L1)-(L2), we have $\set{F}_{\TN{lip}}(\phi) \subset \set{F}_\Phi(\phi)$ where 
$\set{F}_{\TN{lip}}(\phi)$ is the set of $\eta \in \bar{\mathbb{R}}_+^{S\times A}$ satisfying for all $(x',a')$ such that $a' \notin \set{O}(x', \phi)$, 
\begin{align}
\sum_{x \in \set{S}}  \sum_{a \notin \set{O}(x,\phi)}
\eta(x, a) 
\left( \left[ \frac{\delta^*(x', a';\phi)}{H+1} - 2 \left( Ld(x,x')^\alpha
+  L' d(a,a')^{\alpha'} \right)
\right]_+ \right)^2
 \ge 2  \label{eq:ind-r-set}
\end{align}
where we use the notation $[u]_+ :=\max\{0,u\}$ for $u\in \mathbb{R}$.
Furthermore, the optimal values $K_\Phi(\phi)$ and $K_{\TN{lip}}(\phi)$ of $P(\phi, \set{F}_{\Phi}(\phi))$
and $P(\phi, \set{F}_{\TN{lip}}(\phi))$ are upper bounded by $8\frac{(H+1)^3}{\delta^2_{\min}}
 S_{\textnormal{lip}} A_{\textnormal{lip}}$ where 
\begin{align*}
S_{\textnormal{lip}} := 
\min
\left\{S, 
\left( \frac{D \sqrt{d}}{ \left( \frac{\delta_{\min}}{8L  (H+1)} \right)^{1/\alpha}} +1 \right)^{d}
\right\},  
\text{ and }
A_{\textnormal{lip}} := 
\min
\left\{A, 
\left( \frac{D' \sqrt{d'}}{ \left( \frac{\delta_{\min}}{8L'  (H+1)} \right)^{1/\alpha'}} +1 \right)^{d'}
\right\} \;.
\end{align*}
\end{theorem}

The above theorem has important consequences. First, it states that exploiting the Lipschitz structure optimally, one may achieve a regret at most scaling as $\frac{(H+1)^3}{\delta^2_{\min}} S_{\textnormal{lip}} A_{\textnormal{lip}}\log T$. This scaling is independent of the sizes of the state and action spaces provided that the minimal gap $\delta_{\min}$ is fixed, and provided that the span $H$ does not scale with $S$. The latter condition typically holds for fast mixing models or for MDPs with diameter not scaling with $S$ (refer to \cite{regal} for a precise connection between $H$ and the diameter). Hence, exploiting the structure can really yield significant regret improvements. As shown in the next section, leveraging the simplified structure in $\set{F}_{\TN{lip}}(\phi)$, we may devise a simple algorithm achieving these improvements, i.e., having a regret scaling at most as $K_{\TN{lip}}(\phi)\log T$.


\begin{algorithm}[tb]
   \caption{DEL($\gamma$)}
   \label{alg:ORL}
\begin{algorithmic}
\INPUT Model structure $\Phi$
   \STATE Initialize 
   $N_1(x) \leftarrow \mathbbm{1}[x = X_1]$,
   $N_1(x,a) \leftarrow 0$, $s_1(x) \leftarrow 0$, 
$p_1 (y \mid x,a) \leftarrow {1}/{|\set{S}|}$, $r_1 (x,a) \leftarrow 0 $ for each $x,y \in \set{S}$, $ a \in \set{A}$, and $\phi_1$ accordingly.
\FOR{$t=1,...,T$}
\STATE{
For each $x \in \set{S}$, let $\set{C}_t(x) := \{a \in \set{A}  : N_t(x, a) \ge \log^2 N_t(x)\}$, $\phi'_t := \phi_t(\set{C}_t)$, $h'_t(x) := h^*_{\phi'_t}(x)$, $\zeta_t := \frac{1}{1+  \log \log t}$ and
 $\gamma_t := (1+\gamma)(1+ \log t)$}

\IF{$\forall a \in \set{O}(x; \phi'_t)$, $N_t(X_t, a) < \log^2 N_t(X_t) + 1$}
\STATE{{\bf Monotonize: }$A_t \leftarrow {A}^\mnt_t := \argmin_{a \in \set{O}(x; \phi'_t)}  N_t(X_t, a)$.}

\ELSIF{$\exists a \in \set{A}$ s.t. $N_t(X_t, a) < \frac{\log N_t(X_t)}{1+\log \log N_t(X_t)}$}
\STATE{{\bf Estimate: }$A_t \leftarrow {A}^\est_t := \argmin_{a \in \set{A}}  N_t(X_t, a)$.}

\ELSIF{$\left(\frac{N_t (x, a)}{\gamma_t}: (x, a) \in \set{S} \times \set{A} \right) \in \set{F}_{\Phi} (\phi_t; \set{C}_t, \zeta_t)$.
 }
\STATE{{\bf Exploit: }$A_t \leftarrow {A}^\xpt_t:= \argmin_{a \in \set{O}(x; \phi'_t)}  N_t(X_t, a)$.}

\ELSE
\STATE{For each $(x,a) \in \set{S} \times \set{A}$, let $\delta_t(x,a) := \delta^*(x,a; \phi_t, \set{C}_t, \zeta_t)$.
 }
\IF{
$\set{F}_t 
:= \set{F}_{\Phi} (\phi_t; \set{C}_t, \zeta_t)  \cap  \{ \eta :
\eta(x,a) = \infty \text{ if } \delta_t(x,a) = 0 \} = \emptyset$
}
\STATE{Let $\eta_t (x,a) = \infty$ if $\delta_t(x,a) = 0$
and $\eta_t (x,a) = 0$ otherwise.}
\ELSE 
\STATE{Obtain a solution $\eta_t$ of $\set{P}(\delta_t, \set{F}_t)$:
$
\inf_{\eta \in \set{F}_t} \sum_{(x,a) \in \set{S}\times \set{A}} \eta(x, a) \delta_t(x,a)
$
}
\ENDIF 
\STATE{{\bf Explore: }
$A_t \leftarrow {A}^\xpr_t
:= \argmin_{a \in \set{A} : N_t (X_t, a) \le \eta_t(X_t, a) \gamma_t} N_t(X_t,a)$.}

\ENDIF 
      \STATE{Select action $A_t$, and observe the next state $X_{t+1}$ and the instantaneous reward $R_t$. }
          \STATE{ Update $\phi_{t+1}$, $N_{t+1}(x)$ and $N_{t+1}(x, a)$ for each $(x,a) \in \set{S} \times \set{A}$. }      
   \ENDFOR
\end{algorithmic}
\end{algorithm}

\section{Algorithms}
\label{sec:algo}

In this section, we present DEL (Directed Exploration Learning), an algorithm that achieves the regret limits identified in the previous section. Asymptotically optimal algorithms for generic controlled Markov chains have already been proposed in \cite{graves1997}, and could be adapted to our setting. By presenting DEL, we aim at providing simplified, yet optimal algorithms. Moreover, DEL can be adapted so that the exploration rates of sub-optimal actions are {\it directed} towards the solution of an optimization problem $P(\phi,{\cal F}(\phi))$ provided that ${\cal F}(\phi)\subset {\cal F}_\Phi(\phi)$ (it suffices to use $\set{F}(\phi_t)$ instead of $\set{F}_\Phi(\phi_t)$ in DEL). For example, in the case of Lipschitz structure $\Phi$, running DEL on $\set{F}_{\textnormal{lip}} (\phi)$ yields a regret scaling at most as $\frac{(H+1)^3}{\delta^2_{\min}} S_{\textnormal{lip}} A_{\textnormal{lip}}\log T$.

The pseudo-code of DEL with input parameter $\gamma > 0 $ is given in Algorithm~\ref{alg:ORL}. There, for notational convenience, we abuse the notations and  redefine $\log t$ as $ \mathbbm{1} [t \ge 1] \log t$, and let $\infty \cdot 0 = 0$.  
$\phi_t$ refers to the estimated MDP at time $t$ (using empirical transition rates and rewards). For any non-empty correspondence $\set{C} : \set{S} \twoheadrightarrow \set{A}$ (i.e., for any $x$, ${\cal C}(x)$ is a non-empty subset of ${\cal A}$), let $\phi(\set{C})$ denote the restricted MDP where the set of actions available at state $x$ is $\set{C}(x)$. 
Then, $g^*_{\phi(\set{C})}$ and  $h^*_{\phi(\set{C})}$ are the (optimal) gain and bias functions corresponding to the restricted MDP $\phi(\set{C})$. Given a restriction defined by $\set{C}$,
for each $(x,a) \in \set{S}\times \set{A}$, let $\delta^*(x, a; \phi, \set{C}) := (\mathbf{B}^*_{\phi (\set{C})} h^*_{\phi(\set{C})} )(x) -  (\mathbf{B}^a_{\phi} h^*_{\phi(\set{C})}) (x)$
and $H_{\phi(\set{C})} := \max_{x, y\in \set{S}}  h^*_{\phi(\set{C})} (x) -  h^*_{\phi(\set{C})} (y)$.
For $\zeta \ge 0$, let $\delta^*(x, a; \phi, \set{C}, \zeta) : = 0$ if $\delta^*(x, a; \phi, \set{C}) \le \zeta$, and let $\delta^*(x, a; \phi, \set{C}, \zeta) : = \delta^*(x, a; \phi, \set{C})$ otherwise.
For $\zeta \ge 0$, we further define the set of confusing MDPs $\Delta_{\Phi} (\phi;  \set{C}, \zeta)$, and the set of feasible solutions $\set{F}_{\Phi} (\phi;  \set{C}, \zeta)$ as:
$$
\Delta_{\Phi} (\phi;  \set{C}, \zeta)  := 
\left\{\psi \in \Phi \cup \{\phi\} :  
\phi \ll \psi;~
\begin{matrix*}[l]
(i)\ \kl_{\phi \mid \psi}(x,a) = 0 ~\forall x, \forall a \in \set{O}(x; \phi(\set{C}));  \\
(ii)\ \exists (x, a) \in \set{S}\times\set{A} \text{~s.t.~} \\
\qquad
a \notin \set{O}(x;  \phi(\set{C}))~\text{and}~
\delta^*(x,a;\psi, \set{C}, \zeta) = 0
 \end{matrix*}
 \right\}  
 $$
 $$
 \set{F}_{\Phi} (\phi;  \set{C}, \zeta ) := 
\left\{ \eta \in \bar{\mathbb{R}}_{+}^{S \times A}:
 \sum_{x \in \set{S}} \sum_{a \in \set{A}} 
\eta(x,a)
 \kl_{\phi \mid \psi} (x, a)  \ge 1, ~\forall \psi \in \Delta_{\Phi}(\phi; \set{C}, \zeta) 
 \right\} .
$$
Similar sets $\set{F}_{\textnormal{un}} (\phi; \set{C}, \zeta)$ and $\set{F}_{\textnormal{lip}} (\phi; \set{C}, \zeta )$ can be defined for the cases of unstructured and Lipschitz MDPs (refer to the appendix), and DEL can be simplified in these cases by replacing $ \set{F}_{\Phi} (\phi;  \set{C},\zeta)$ by $\set{F}_{\textnormal{un}} (\phi;  \set{C},\zeta)$ or $\set{F}_{\textnormal{lip}} (\phi;  \set{C},\zeta)$ in the pseudo-code. Finally, $\set{P}(\delta, \set{F})$ refers to the optimization problem $\inf_{\eta \in \set{F}} \sum_{(x,a) \in \set{S}\times \set{A}} \eta(x, a) \delta(x,a)$.

DEL combines the ideas behind OSSB \cite{combesnips17}, an asymptotically optimal algorithm for structured bandits, and the asymptotically optimal algorithm presented in \cite{burnetas1997optimal} for RL problems without structure. DEL design aims at exploring sub-optimal actions no more than what the regret lower bound prescribes. To this aim, it essentially solves in each iteration $t$ an optimization problem close to $P(\phi_t,\set{F}_\Phi(\phi_t))$ where $\phi_t$ is an estimate of the true MDP $\phi$. Depending on the solution and the number of times apparently sub-optimal actions have been played, DEL decides to explore or exploit. The estimation phase ensures that certainty equivalence holds. The "monotonization" phase together with the restriction to relatively well selected actions were already proposed in \cite{burnetas1997optimal} to make sure that accurately estimated actions only are selected in the exploitation phase. The various details and complications introduced in DEL ensure that its regret analysis can be conducted. In practice (see the appendix), our initial experiments suggest that many details can be removed without large regret penalties.   

\begin{theorem} \label{thm:DEL}
For a structure $\Phi$ with Bernoulli rewards and for any ergodic MDP $\phi \in \Phi$,
assume that:
($i$) $\phi$ is in the interior of $\Phi$ (i.e., 
there exists a constant $\zeta_0 >0$ such that for any $\zeta \in (0, \zeta_0)$,
$\psi \in \Phi$ if $\|\phi - \psi\| \le \zeta$ and $\psi \ll \phi$), 
($ii$) the solution $\eta^*(\phi)$ is uniquely defined for each $(x,a)$ such that $a  \notin \set{O} (x;\phi)$,
($iii$) continuous at $\phi$ (i.e., for any given $\varepsilon > 0$, there exists $\zeta(\varepsilon) > 0$ such that for all $\zeta \in (0, \zeta(\varepsilon))$, if $\|\psi - \phi\| \le \zeta$, $\max_{(x,a) : a \notin \set{O}(x; \phi) } |\eta^*(x,a; \psi, \zeta) - \eta^*(x,a; \phi)|  \le \varepsilon$ where $\eta^*(\psi, \zeta)$ is solution of $\set{P}(\delta^*(\psi, \set{A}, \zeta), \set{F}_\Phi(\psi; \set{A}, \zeta))$, and $\eta^*(x,a; \phi)$ that of $P(\phi,\set{F}_\Phi(\phi))$).
Then, for $\pi = \textnormal{DEL($\gamma$)}$ with any $\gamma  > 0$, we have:
\begin{align} \label{eq:upper}
\limsup_{T \to \infty}\frac{R^\pi_T(\phi)}{\log T} \le (1+\gamma) K_{\Phi}(\phi)  \;.
\end{align}

For Lipschitz $\Phi$ with (L1)-(L2) (resp. unstructured $\Phi$), if $\pi=\textnormal{DEL}$ uses in each step $t$, $\set{F}_{\textnormal{lip}} (\phi_t; \set{C}_t, \zeta_t)$ (resp. $ \set{F}_{\textnormal{un}} (\phi_t; \set{C}_t, \zeta_t)$) instead of $\set{F}_{\Phi} (\phi_t; \set{C}_t, \zeta_t)$, its regret is asymptotically smaller than $(1+\gamma)K_{\textnormal{lip}}(\phi)\log T$ (resp. $(1+\gamma)K_{\textnormal{un}}(\phi)\log T$).
\end{theorem}

In the above theorem, the assumptions about the uniqueness and continuity of the solution $\eta^*(\phi)$ could be verified for particular structures. In particular, we believe that they generally hold in the case of unstructured and Lipschitz MDPs. Also note that similar assumptions have been made in \cite{graves1997}. 




\section{Extensions and Future Work}

It is worth extending the approach developed in this paper to the case of structured discounted RL problems (although for such problems, there is no ideal way of defining the regret of an algorithm). There are other extensions worth investigating. For example, since our framework allows any kind of structure, we may specify our regret lower bounds for structures stronger than that corresponding to Lipschitz continuity, e.g., the reward may exhibit some kind of unimodality or convexity. Under such structures, the regret improvements might become even more significant. Another interesting direction consists in generalizing the results to the case of communicating MDPs. This would allow us for example to consider deterministic system dynamics and unknown probabilistic rewards. 


\section{Acknowledgements} 
This work was partially supported by the Wallenberg AI, Autonomous Systems and Software Program (WASP) funded by the Knut and Alice Wallenberg Foundation. Jungseul Ok is now with UIUC in Prof. Sewoong Oh's group. He would like to thank UIUC for financially supporting his participation to NIPS 2018 conference.  

\vfill
\newpage
{\small
\bibliographystyle{plainnat}
\bibliography{ref,vr2017}
}
\clearpage
\appendix


\section{The DEL Algorithm}

In this section, we present DEL, and state its asymptotic performance guarantees. DEL pseudo-code is given in Algorithm \ref{alg:ORL}.
There, for notational convenience, we abuse the notations and  redefine $\log t$ as $ \mathbbm{1} [t \ge 1] \log t$.  
$\phi_t$ refers to the estimated MDP at time $t$ (e.g. using empirical transition rates). For non-empty correspondence $\set{C} : \set{S} \twoheadrightarrow \set{A}$ (i.e., for any $x$, ${\cal C}(x)$ is a non-empty subset of ${\cal A}$), let $\phi(\set{C})$ denote the restricted MDP where the set of actions available at state $x$ is limited to $\set{C}(x)$. 
Then, $g^*_{\phi(\set{C})}$ and  $h^*_{\phi(\set{C})}$ are the (optimal) gain and bias functions corresponding to the restricted MDP $\phi(\set{C})$, respectively. Given a restriction defined by $\set{C}$,
for each $(x,a) \in \set{S}\times \set{A}$, let $\delta^*(x, a; \phi, \set{C}) := (\mathbf{B}^*_{\phi (\set{C})} h^*_{\phi(\set{C})} )(x) -  (\mathbf{B}^a_{\phi} h^*_{\phi(\set{C})}) (x)$
and $H_{\phi(\set{C})} := \max_{x, y\in \set{S}}  h^*_{\phi(\set{C})} (x) -  h^*_{\phi(\set{C})} (y)$.
For $\zeta \ge 0$, let $\delta^*(x, a; \phi, \set{C}, \zeta) : = 0$ if $\delta^*(x, a; \phi, \set{C}) \le \zeta$, and let $\delta^*(x, a; \phi, \set{C}, \zeta) : = \delta^*(x, a; \phi, \set{C})$ otherwise.
For $\zeta \ge 0$, we further define the set of confusing MDPs $\Delta_{\Phi} (\phi;  \set{C}, \zeta)$, and the set of feasible solutions $\set{F}_{\Phi} (\phi;  \set{C}, \zeta)$:
$$
\Delta_{\Phi} (\phi;  \set{C}, \zeta)  := 
\left\{\psi \in \Phi \cup \{\phi\} :  
\phi \ll \psi;~
\begin{matrix*}[l]
(i)\ \kl_{\phi \mid \psi}(x,a) = 0 ~\forall x, \forall a \in \set{O}(x; \phi(\set{C}));  \\
(ii)\ \exists (x, a) \in \set{S}\times\set{A} \text{~s.t.~} \\
\qquad
a \notin \set{O}(x;  \phi(\set{C}))~\text{and}~
\delta^*(x,a;\psi, \set{C}, \zeta) = 0
 \end{matrix*}
 \right\}  
 $$
 $$
 \set{F}_{\Phi} (\phi;  \set{C}, \zeta ) := 
\left\{ \eta \in \bar{\mathbb{R}}_{+}^{S \times A}:
 \sum_{x \in \set{S}} \sum_{a \in \set{A}} 
\eta(x,a)
 \kl_{\phi \mid \psi} (x, a)  \ge 1 ~\forall \psi \in \Delta_{\Phi}(\phi; \set{C}, \zeta) 
 \right\} .
$$

For the unstructured and Lipschitz MDPs, we simplify
the feasible solution set
as $\set{F}_{\textnormal{un}} (\phi; \set{C}, \zeta)$ and 
$\set{F}_{\textnormal{lip}} (\phi; \set{C}, \zeta )$
, respectively, defined as:
\begin{align*}
\set{F}_{\textnormal{un}} (\phi;  \set{C}, \zeta) := 
\left\{ \eta \in \bar{\mathbb{R}}_{+}^{S \times A}:
\eta(x,a) \left( \frac{\delta^* (x,a; \phi, \set{C}, \zeta)}{H_{\phi(\set{C})} + 1}\right)^2  \ge 2 ~~\forall 
(x,a) ~\text{s.t.}~ a \notin \set{O} (x ; \phi(\set{C}))
 \right\} 
\end{align*}
\begin{align*}
 \set{F}_{\textnormal{lip}} (\phi;  \set{C}, \zeta ) := 
\Bigg\{ \eta \in \bar{\mathbb{R}}_{+}^{S \times A}:
 L_{\textnormal{lip}} (x', a' ; \phi, \set{C}, \zeta) \ge 2 \forall (x',a') ~\text{s.t.}~ a' \notin \set{O} (x' ; \phi(\set{C}))\Bigg\}
\end{align*}
where
\begin{align*}
L_{\textnormal{lip}} (x', a' ; \phi, \set{C}, \zeta) :=  
\sum_{x \in \set{S}} \sum_{a \in \set{A}}
\eta(x,a) 
\left( \bigg[ \tfrac{\delta^* (x',a'; \phi, \set{C}, \zeta)}{H_{\phi(\set{C})} + 1}
 - 2\left(Ld(x,x')^\alpha +L' d(a, a')^{\alpha'} \right)
\bigg]_+ \right)^2. 
\end{align*}

\begin{algorithm}[!tb]
   \caption{DEL($\gamma$)}
   \label{alg:ORL}
\begin{algorithmic}
\INPUT Model structure $\Phi$
   \STATE Initialize 
   $N_1(x) \leftarrow \mathbbm{1}[x = X_1]$,
   $N_1(x,a) \leftarrow 0$, $s_1(x) \leftarrow 0$, 
$p_1 (y \mid x,a) \leftarrow {1}/{|\set{S}|}$, $r_1 (x,a) \leftarrow 0 $ for each $x,y \in \set{S}$, $ a \in \set{A}$, and $\phi_1$ accordingly.
\FOR{$t=1,...,T$}
\STATE{
For each $x \in \set{S}$, let $\set{C}_t(x) := \{a \in \set{A}  : N_t(x, a) \ge \log^2 N_t(x)\}$, $\phi'_t := \phi_t(\set{C}_t)$, $h'_t(x) := h^*_{\phi'_t}(x)$, $\zeta_t := \frac{1}{1+  \log \log t}$ and
 $\gamma_t := (1+\gamma)(1+ \log t)$}

\IF{$\forall a \in \set{O}(x; \phi'_t)$, $N_t(X_t, a) < \log^2 N_t(X_t) + 1$}
\STATE{{\bf Monotonize: }$A_t \leftarrow {A}^\mnt_t := \argmin_{a \in \set{O}(x; \phi'_t)}  N_t(X_t, a)$.}

\ELSIF{$\exists a \in \set{A}$ s.t. $N_t(X_t, a) < \frac{\log N_t(X_t)}{1+\log \log N_t(X_t)}$}
\STATE{{\bf Estimate: }$A_t \leftarrow {A}^\est_t := \argmin_{a \in \set{A}}  N_t(X_t, a)$.}

\ELSIF{$\left(\frac{N_t (x, a)}{\gamma_t}: (x, a) \in \set{S} \times \set{A} \right) \in \set{F}_{\Phi} (\phi_t; \set{C}_t, \zeta_t)$.
 }
\STATE{{\bf Exploit: }$A_t \leftarrow {A}^\xpt_t:= \argmin_{a \in \set{O}(x; \phi'_t)}  N_t(X_t, a)$.}

\ELSE
\STATE{For each $(x,a) \in \set{S} \times \set{A}$, let $\delta_t(x,a) := \delta^*(x,a; \phi_t, \set{C}_t, \zeta_t)$.
 }
\IF{
$\set{F}_t 
:= \set{F}_{\Phi} (\phi_t; \set{C}_t, \zeta_t)  \cap  \{ \eta :
\eta(x,a) = \infty \text{ if } \delta_t(x,a) = 0 \} = \emptyset$
}

\STATE{Let $\eta_t (x,a) = \infty$ if $\delta_t(x,a) = 0$
and $\eta_t (x,a) = 0$ otherwise.}

\ELSE 
\STATE{Obtain a solution $\eta_t$ of $\set{P}(\delta_t, \set{F}_t)$:
$
\inf_{\eta \in \set{F}_t} \sum_{(x,a) \in \set{S}\times \set{A}} \eta(x, a) \delta_t(x,a)
$
}
\ENDIF 
\STATE{{\bf Explore: }
$A_t \leftarrow {A}^\xpr_t
:= \argmin_{a \in \set{A} : N_t (X_t, a) \le \eta_t(X_t, a) \gamma_t} N_t(X_t,a)$.}

\ENDIF 
      \STATE{Select action $A_t$, and observe the next state $X_{t+1}$ and the instantaneous reward $R_t$. }
          \STATE{ Update $\phi_{t+1}$: 
          
          \begin{minipage}{0.7\textwidth}
    \begin{align*}
       &N_{t+1}(x) \leftarrow  N_t(x) + \mathbbm{1}[x = X_{t+1}], \\ 
     &N_{t+1}(x,a) \leftarrow  N_t(x,a) + \mathbbm{1}[(x,a) = (X_t, A_t)],\\
    &p_{t+1}(y \mid x,a) \leftarrow 
    \begin{cases}
    \frac{N_{t}(x, a) p_{t}(y \mid x,a) + \mathbbm{1}[y=X_{t+1}]}{N_{t+1}(x, a)}  & \text{if $(x,a) = (X_t, A_t)$} \\
        p_{t}(y \mid x,a) & \text{otherwise}
	\end{cases}  \\
    &r_{t+1}(x,a) 
    \leftarrow 
    \begin{cases}
    \frac{N_{t}(x, a) r_{t}(x,a) + R_t}{N_{t+1}(x, a)}  & \text{if $(x,a) = (X_t, A_t)$} \\
        r_{t}(x,a) & \text{otherwise}
	\end{cases} 
\end{align*}
          \end{minipage},
           $ \quad \forall x,y \in \set{S}, a \in \set{A}$
   }      
   \ENDFOR
\end{algorithmic}
\end{algorithm}

\section{Numerical Experiments}

In this section, we briefly illustrate the performance of a simplified version of the DEL algorithm on a simple example constructed so as to comply to a Lipschitz structure. Our objective is to investigate the regret gains obtained by exploiting a Lipschitz structure, and we compare the performance of our two simplified versions of DEL with $\gamma = 1$ and $\zeta_t = 0$, one solving ${P}(\phi_t, \set{F}_{\textnormal{un}} (\phi_t; \set{C}_t, \zeta_t))$ in step $t$, and the other solving ${P}(\phi_t, \set{F}_{\textnormal{lip}} (\phi_t; \set{C}_t, \zeta_t))$.


{\bf The RL problem.} We consider a toy MDP whose states are partitioned into two {\em clusters} $\set{S}_1$, $\set{S}_2$ of equal sizes $S/2$. Both states and actions are embedded into $\mathbb{R}$: 
\begin{itemize}[leftmargin=0.2in]
\item The states in cluster $\set{S}_1$ (resp. $\set{S}_2$) are randomly generated in the interval $[-\zeta,0]$ (resp. $[1,1+\zeta]$) for some $\zeta\in (0,1)$;
\item In each state there are two possible actions: $s=0$ (stands for {\it stay}) and $m=1$ (stands for {\it move}).
\end{itemize}
The transition probabilities depend on the states only through their corresponding clusters, and are given by: for $\epsilon \in (0, 0.5)$, 
\begin{align}
p(y|x,a) = \begin{cases}
\frac{2(1-\epsilon)}{S} & \quad \text{if }(x,y,a)\in \Gamma_p\\
\frac{2\epsilon}{S} & \quad \text{otherwise}
\end{cases},
\end{align}
where 
\begin{align*}
\Gamma_p:=\{(x,y,a): a=s,~\exists i\in\{1,2\}, x,y\in \set{S}_i\} \cup \{(x,y,a): a=m,~\exists i\in\{1,2\}, x\in \set{S}_i, y\notin \set{S}_i\}.
\end{align*}
In words, when the agent decides to move, she will end up in a state uniformly sampled from the other cluster with probability $1-\epsilon$; when she decides to stay, she changes state within her cluster uniformly at random. We take $\epsilon>0$ to ensure irreducibility. For numerical experiments we take $\epsilon=0.1$ and $\zeta = 0.1$.
The reward is obtained according to the following deterministic rule:
\begin{align}
r(x,a) = \begin{cases}
1 \quad \text{if }(x,a): a=m \text{ and }x\in \set{S}_1,\\
0 \quad \text{otherwise.}
\end{cases}
\end{align}
A reward is collected when the agent is in cluster $\set{S}_1$ and decides to move. The optimal stationary strategy consists in moving in each state.

\begin{figure}[!h]
\centering
\begin{subfigure}{0.48\textwidth} 
	\includegraphics[width=0.99\columnwidth]{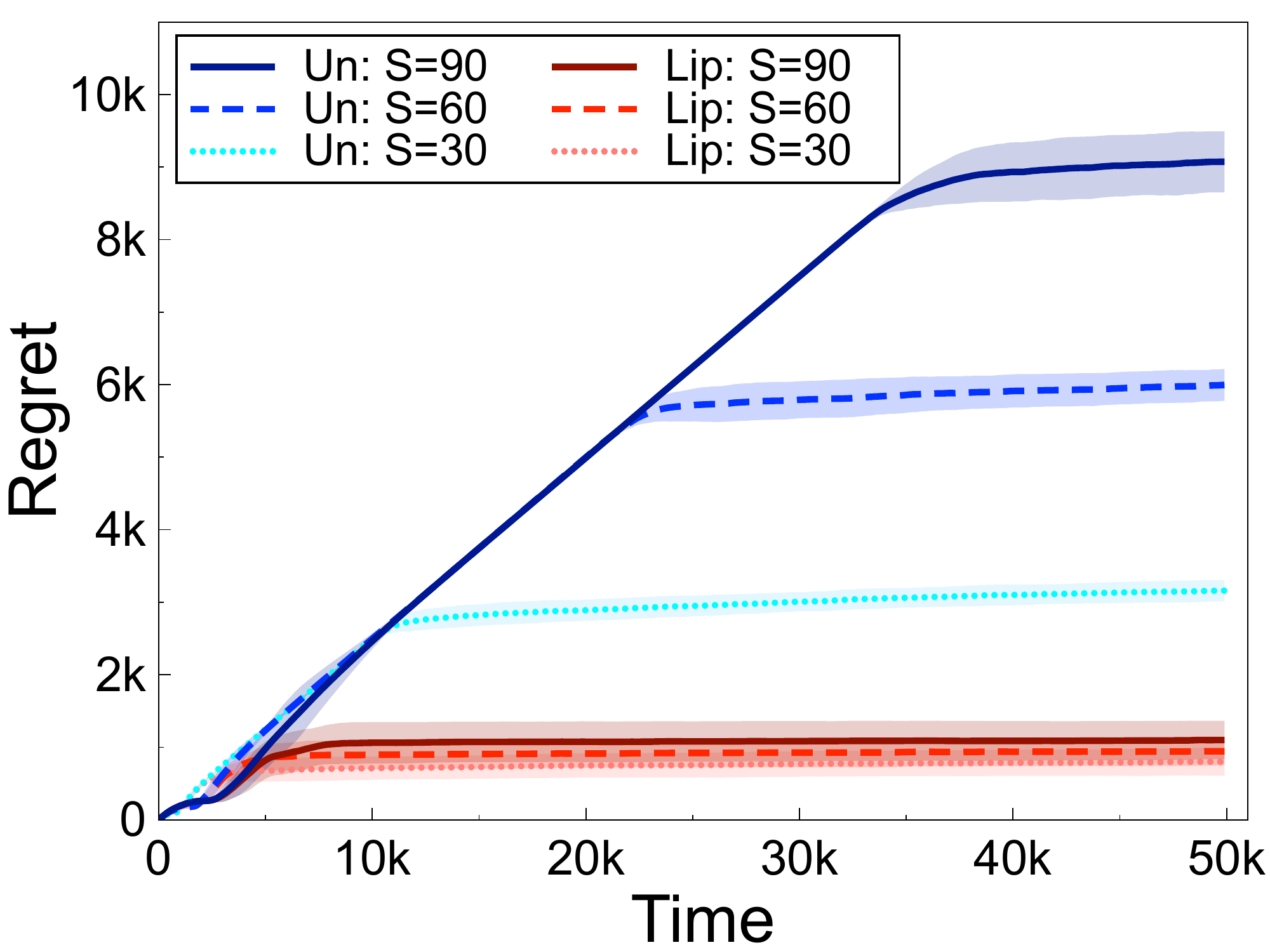} 
		\caption{Regret over time} \label{fig:individual}
\end{subfigure} \hspace{0.2cm}
\begin{subfigure}{0.48\textwidth} 
	\includegraphics[width=0.99\columnwidth]{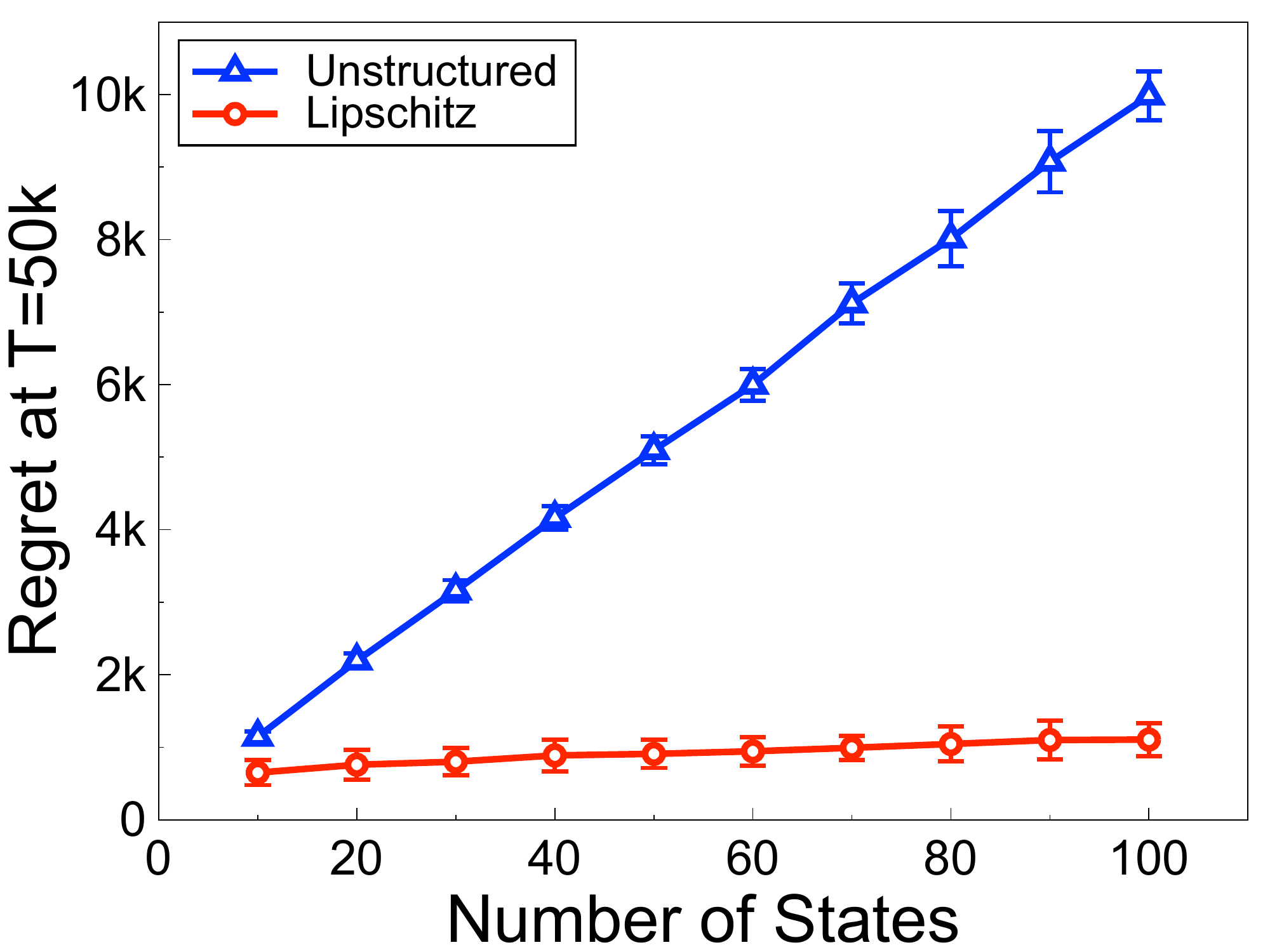}
\caption{Regret at $T = 50$k varying $S$} \label{fig:total}
\end{subfigure}
%

	\caption{Averaged regret under the two simplified versions of DEL over $48$ random samples: {\tt Unstructured} (or {\tt Un}) and {\tt Lipchitz} (or {\tt Lip}) refer to the algorithm with $\set{F}_{\textnormal{un}}$ 
	and $\set{F}_{\textnormal{lip}}$, respectively. The shadows and error bars show one standard deviation.}
	\label{fig:compare}

\end{figure}

Figure~\ref{fig:compare} presents the regret of the two versions of our DEL algorithm. Clearly, exploiting the structure brings a very significant performance improvement and the gain grows as the number of states increases, as predicted by our theoretical results. Observe that the regret after $T=50$k steps under the version of DEL exploiting the Lipschitz structure barely grows with the number of states, see Figure~\ref{fig:compare}(\subref{fig:total}), which was also expected.

\section{Proof of Theorem~\ref{thm:g-r-lower}}

{\bf Notations and preliminaries.} Let $N_T (x) = \sum_{t=1}^{T}\mathbbm{1}[X_t = x]$  and $N_T (x,a) = \sum_{t=1}^{T}\mathbbm{1}[X_t = x, A_t = a]$ denote the number of times $x$ and $(x,a)$ have been visited up to step $T$. For any $\psi\in \Phi$ and any initial state $x_1$, we denote by $\Pr^\pi_{\psi \mid x_1}$ and $\EXP^\pi_{\psi \mid x_1}$ the probability measure and expectation under $\pi$ and $\psi$ conditioned on $X_1 = x_1$. The regret up to step $T$ starting in state $x_1$ under $\pi$ and $\psi$ is denoted by $R_{T,\psi}^\pi(x_1)$. To emphasize the dependence on the MDP $\psi$ of the gap function $\delta^\star$, we further denote its value at $(x,a)$ by $\delta^\star(x,a;\psi)$.

For any $\psi\in \Phi$, using the ergodicity of $\psi$, we may leverage the same arguments as those used in the proof of Proposition~1 of \cite{burnetas1997optimal} to establish a connection between the regret of an algorithm $\pi\in \Pi$ under $\psi$ and $N_T (x,a)$. Specifically, for any $x_1$,
\begin{align} \label{eq:translation}
R^\pi_{T,\psi} (x_1) =  \sum_{x \in \set{S}} \sum_{a \notin \set{O}(y, \psi)} \EXP^\pi_{\psi\mid x_1}[ N_T(x, a)] \delta^\star(x,a;\psi)  + O(1) \;, \quad \text{as $T \to \infty$} \;.
\end{align}

In addition, due to the ergodicity of $\psi$, we can also prove as in Proposition 2 in \cite{burnetas1997optimal} that there exists constants $C, \rho >0$ such that for any $x \in \set{S}$, $\pi \in \Pi$,
\begin{align} \label{eq:erg}
\Pr^\pi_{\psi\mid x}[N_T (x) \le \rho T] \le C \cdot \exp (-\rho T/2).
\end{align}

{\bf Change-of-measure argument.} Let $\pi$ be a uniformly good algorithm, and $x_1$ an initial state. For any {\it bad} MDP $\psi \in \Delta_\Phi (\phi)$, the argument consists in (i) relating the log-likelihood of the observations under $\phi$ and $\psi$ to the expected number of times sub-optimal actions are selected under $\pi$, and (ii) using the fact that $\pi$ is uniformly good to derive a lower bound on the log-likelihood. 

(i) Define by $L$ the log-likelihood of the observations up to step $T$ under $\phi$ and $\psi$. We can use the same techniques as in \cite{kaufmann2016complexity, garivier193} (essentially an extension of Wald's lemma):
\begin{equation}\label{eq:log1}
\mathbb{E}_{\phi\mid x_1}^\pi[L]=\sum_{x,a} \mathbb{E}_{\phi\mid x_1}^\pi[N_T(x,a)] \kl_{\phi \mid \psi}(x,a).
\end{equation}
The so-called data processing inequality \cite{garivier193} yields for all event $\set{E}$ in $\set{H}_T^\pi$: \\
$\mathbb{E}_{\phi\mid x_1}^\pi[L]\ge kl(\Pr^\pi_{\phi \mid x_1}[ {\set{E}}], \Pr^\pi_{\psi \mid x_1}[ {\set{E}}])$, where for $u,v\in [0,1]$, $kl(u,v ) := u\log \frac{u}{v} + (1-u) \log \frac{1-u}{1-v}$. Combine with (\ref{eq:log1}), this leads to:
\begin{equation}\label{eq:log2}
\sum_{x,a\notin \set{O}(x,\phi)} \mathbb{E}_{\phi\mid x_1}^\pi[N_T(x,a)] \kl_{\phi \mid \psi}(x,a)\ge  \kl(\Pr^\pi_{\phi \mid x_1}[ {\set{E}}], \Pr^\pi_{\psi \mid x_1}[ {\set{E}}]).
\end{equation}
Note that in the above sum, we removed $a\in \set{O}(x,\phi)$ since $\kl_{\phi \mid \psi}(x,a) = 0$ if $a \in \set{O}(x,\phi)$.

(ii) Next we will leverage the fact that $\pi$ is uniformly good to select the event $E$. We first state the following lemma, proved at the end of this section. Since $\Pi^\star(\phi)\cap \Pi^\star(\psi)=\emptyset$, there exists $x \in \set{S}$ such that for all $ \alpha  >0$,
$$
\EXP^\pi_{\psi \mid x_1}\left[ \sum_{a \in \set{O}(x, \phi)} N_T (x, a)\right] = o(T^\alpha).
$$
Indeed, otherwise $\pi$ would not be uniformly good. Now define the event $\set{E}$ as:
$$
\set{E} := \left[
N_T(x) \ge \rho T,  
 \sum_{a \notin \set{O}(x, \phi)}  N_T (x, a) \le  \sqrt{T} \right],
$$
where the constant $\rho$ is chosen so that (\ref{eq:erg}) holds under $\phi$ and $\psi$. Using a union bound, we have
\begin{align}
1- \Pr^\pi_{\phi \mid x_1} [\set{E}] 
&\le
\Pr^\pi_{\phi \mid x_1} \left[
N_T(x) \le \rho T \right]
+
\Pr^\pi_{\phi \mid x_1} \left[
 \sum_{a \notin \set{O}(x, \phi)}  N_T (x, a) \ge  \sqrt{T} \right] \nonumber \\
 & \le
C \cdot \exp(- \rho T /2)
+
\frac{\EXP^\pi_{\phi \mid x_1} \left[\sum_{a \notin \set{O}(x, \phi) }  N_T (x, a) \right]}{ \sqrt{T}} \label{eq:event-in-phi}
\end{align}
where for the first and second terms in the last inequality, we used (\ref{eq:erg}) and Markov inequality, respectively.
Since $\pi$ is uniformly good, $\EXP^\pi_{\phi \mid x_1} [\sum_{a \notin \set{O}(x, \phi)}N_T(x,a)] = o(T^\alpha)$ for all $\alpha > 0$, the last term of \eqref{eq:event-in-phi} converges to $0$, i.e.,
$\Pr^\pi_{\phi \mid x_1}[\set{E}] \to 1$ as $T \to \infty$.
Using Markov inequality, it follows that
\begin{align*}
\Pr^\pi_{\psi \mid x_1} [\set{E}] 
&\le
\Pr^\pi_{\psi \mid x_1}
\left[
N_T(x) -
 \sum_{a \notin \set{O}(x, \phi)}  N_T (x, a) \ge 
\rho T - \sqrt{T} \right]\\
&\le
\frac{\EXP^\pi_{\psi \mid x_1} [
N_T(x) - \sum_{a \notin \set{O}(x, \phi)} N_T(x, a) ]}{\rho  T - \sqrt{T}} 
=
\frac{\EXP^\pi_{\psi \mid x_1} [
\sum_{a \in \set{O}(x, \phi)} N_T(x, a) ]}{\rho  T - \sqrt{T}} 
\end{align*}
which converges to $0$ because of our choice of $x$. Combining $\Pr^\pi_{\phi \mid x_1}[\set{E}] \to 1$
and $\Pr^\pi_{\psi \mid x_1}[\set{E}] \to 0$,
\begin{align*}
\frac{kl( \Pr^\pi_{\phi \mid x_1} [\set{E}], \Pr^\pi_{\psi \mid x_1} [\set{E}])}{\log T} 
\underset{T \to \infty}{\sim}
\frac{1}{\log T} \log \left(\frac{1}{\Pr^\pi_{\psi \mid x_1}[\set{E}]}  \right)
\ge
\frac{1}{\log T}
\log 
\left(
\frac{\rho  T - \sqrt{T}}{\EXP^\pi_{\psi \mid x_1} [\sum_{a \in \set{O}(x, \phi)} N_T(x, a) ]}
\right)
\end{align*}
which converges to $1$ as $T$ grows large due to our choice of $x$. Plugging this result in (\ref{eq:log2}), we get:
\begin{equation}\label{eq:log3}
\lim\inf_{T\to\infty}{1\over \log T} \sum_{x,a\notin \set{O}(x,\phi)} \mathbb{E}_{\phi\mid x_1}^\pi[N_T(x,a)] \kl_{\phi \mid \psi}(x,a)\ge  1.
\end{equation}
Combining the above constraints valid for any $\psi\in  \Delta_\Phi (\psi)$ and (\ref{eq:translation}) concludes the proof of the theorem.


\section{Proof of Theorem~\ref{thm:unrelated}}

We first prove the decoupling lemma.

{\bf Proof of Lemma~\ref{lem:gen-decom}}. 
We prove the lemma by contradiction. Assume that 
  $\Pi^\star(\phi) \cap \Pi^\star(\psi_1) \neq \emptyset$ and
  $\Pi^\star(\phi) \cap \Pi^\star(\psi_2) \neq \emptyset$. 
  Let $\Pi^\star(\phi, \psi_1, \psi_2) :=  \Pi^\star(\phi) \cap \Pi^\star(\psi_1) \cap \Pi^\star(\psi_2)$.
It is sufficient to show  
\begin{align}
(i)~ \Pi^\star(\phi, \psi_1, \psi_2) \neq \emptyset  \;, \quad \text{and} \quad
(ii)~ \Pi^\star(\phi, \psi_1, \psi_2) \subseteq \Pi^\star(\psi_0)  \;.
\label{eq:decom-wts}
\end{align}
Indeed, this implies $\Pi^\star(\phi) \cap \Pi^\star(\psi_0) \neq \emptyset$.
Note that any policy $f \in \Pi^*(\phi)$ has the same gain and bias function under $\phi, \psi_0, \psi_1, \psi_2$ since the modifications of $\phi$ are made on suboptimal (state, action) pairs. Specifically,
\begin{align} \label{eq:identical}
g^f_{\psi_0} =  g^f_{\psi_1} = g^f_{\psi_2}=g^\star_\phi   \quad \text{and} \quad 
h^f_{\psi_0} =  h^f_{\psi_1} = h^f_{\psi_2}=h^\star_\phi  \;.
\end{align}

To prove (i), the first part of \eqref{eq:decom-wts}, 
consider a policy $f' \in \Pi^\star(\phi) \cap \Pi^\star(\psi_1)$ and a policy $f'' \in \Pi^\star(\phi) \cap \Pi^\star(\psi_2)$.
Then, from the optimality of $f'$ under $\psi_1$, it follows that 
for each $x \in \set{S}$,
\begin{align} \label{eq:one-side-ineq}
g^\star_{\psi_1} = g^{f'}_{\psi_1} \ge g^{f''}_{\psi_1} = g^{f''}_{\psi_2}= g^{\star}_{\psi_2}\;, 
\end{align}
where for the second equality, we use \eqref{eq:identical}.
Similarly, we have for each $x \in \set{S}$,
\begin{align*}
g^\star_{\psi_2} = g^{f''}_{\psi_2} \ge g^{f'}_{\psi_2} = g^{f'}_{\psi_1}= g^{\star}_{\psi_1} \;.
\end{align*}
Hence $g^\star_{\psi_1} = g^\star_{\psi_2}$ and $\Pi^\star(\phi, \psi_1, \psi_2) = \Pi^\star(\phi) \cap \Pi^\star(\psi_1) = \Pi^\star(\phi) \cap \Pi^\star(\psi_2) \neq \emptyset$.

We now prove (ii), the second part of \eqref{eq:decom-wts}. Let $f\in \Pi^\star(\phi, \psi_1, \psi_2)$. It is sufficient to show $g^f_{\psi_0}$ and $h^f_{\psi_0}(x)$ verify the Bellman optimality equation for model $\psi_0$. 
Using \eqref{eq:identical} and the optimality of $f$ under $\psi_1$, for all $a\in \set{A}$, if $(x,a) \notin \set{U}_2$, 
\begin{align} 
r_{\psi_0} (x, f(x)) + \sum_{y \in \set{S}} p_{\psi_0}(y | x, f(x)) h^f_{\psi_0} (y) 
 & \overset{(a)}{=} r_{\psi_1} (x, f(x)) + \sum_{y \in \set{S}} p_{\psi_1}(y | x, f(x)) h^f_{\psi_1} (y) \nonumber \\
&\overset{(b)}{\ge}
r_{\psi_1} (x, a) + \sum_{y \in \set{S}} p_{\psi_1}(y | x, a) h^f_{\psi_1} (y) \nonumber \\
& \overset{(c)}{=} r_{\psi_0} (x, a) + \sum_{y \in \set{S}} p_{\psi_0}(y | x, a) h^f_{\psi_0} (y) \;,
\label{eq:BO-S'}
\end{align} 
where for (a) and (c), we used \eqref{eq:identical} and the fact that the kernels of $\psi_0$ and $\psi_1$
are the same at every $(x,a) \notin \set{U}_2$,
and for (b), we used the fact that $g^f_{\psi_1}$ and $h^f_{\psi_1}$ verify the Bellman optimality equation for $\psi_1$.
Similarly, using the optimality of $f$ under $\psi_2$, it follows that 
for $(x,a) \in \set{U}_2$, 
\begin{align} 
r_{\psi_0} (x, f(x)) + \sum_{y \in \set{S}} p_{\psi_0}(y | x, f(x)) h^f_{\psi_0} (y) 
 & = r_{\psi_2} (x, f(x)) + \sum_{y \in \set{S}} p_{\psi_2}(y | x, f(x)) h^f_{\psi_2} (y) \nonumber \\
& \ge
r_{\psi_2} (x, a) + \sum_{y \in \set{S}} p_{\psi_2}(y | x, a) h^f_{\psi_2} (y) \nonumber  \\
& = r_{\psi_0} (x, a) + \sum_{y \in \set{S}} p_{\psi_0}(y | x, a) h^f_{\psi_0} (y) \;. \label{eq:BO-S''}
\end{align} 
Combining \eqref{eq:BO-S'} and \eqref{eq:BO-S''}, for all $(x,a) \in \set{S} \times \set{A}$,
\begin{align*} 
r_{\psi_0} (x, f(x)) + \sum_{y \in \set{S}} p_{\psi_0}(y | x, f(x)) h^f_{\psi_0} (y) 
&\ge
r_{\psi_0} (x, a) + \sum_{y \in \set{S}} p_{\psi_0}(y | x, a) h^f_{\psi_0} (y) \;,
\end{align*} 
which implies that 
$g^f_{\psi_0}$ and $h^f_{\psi_0}$ verify the Bellman optimality equation under model $\psi_0$, i.e., $f\in \Pi^*(\psi_0)$.
\ep

{\bf Proof of Theorem~\ref{thm:unrelated}}. 
Recall that any policy $f \in \Pi^*(\phi)$ has the same gain and bias function in $\psi$ and $\phi$
since the kernels of $\phi$ and $\psi$ are identical at every $(x,a)$ such that $a \in \set{O}(x; \phi)$.
More formally, for any $f \in \Pi^*(\phi)$,
\begin{align*} 
\mathbf{B}^f_\phi = \mathbf{B}^f_\psi, \quad 
g^*_\phi = g^f_\phi =  g^f_{\psi} \quad \text{and} \quad 
h^*_\phi (\cdot) = h^f_\phi (\cdot) =  h^f_{\psi}(\cdot)  \;.
\end{align*}

Next we show that for all $\psi \in \Delta_\Phi(\phi)$,
\begin{align}
\Pi^*(\phi) \cap\Pi^*(\psi) = \emptyset
\implies
\exists (x,a)~\text{such that}~
(\mathbf{B}^{a}_\psi h^*_\phi)(x) >
g^*_\phi + h^*_\phi(x) 
\;. \label{eq:opt2}
\end{align}

We prove (\ref{eq:opt2}) by contradiction. Consider a policy $f \in \Pi^*(\phi)$.
Suppose that for all $(x,a)$, $(\mathbf{B}^{a}_\psi h^f_\phi)(x) \le
 g^f_\phi + h^f_\phi(x)$. Then, for all $(x,a)$,
\begin{align*}
(\mathbf{B}^{f}_\psi h^f_\psi)(x) = (\mathbf{B}^{f}_\phi h^f_\phi)(x)
=
g^f_\phi + h^f_\phi(x)  \ge 
\max_{a \in \set{A}} (\mathbf{B}^a_\psi h^f_\phi) (x)
\end{align*}
which implies that
 $g^f_\psi$  and $h^f_\psi $ verify the Bellman optimality equation under $\psi$.
 Hence, $f \in\Pi^*(\psi)$ which contradicts to $\Pi^*(\phi) \cap \Pi^*(\psi) = \emptyset$.
 
 Finally Theorem~\ref{thm:unrelated} is obtained by combining the decoupling lemma and (\ref{eq:opt2}). Indeed, due to the decoupling lemma, we may restrict $\Delta_\Phi(\phi)$ to MDPs obtained from $\phi$ by only changing the kernels in a single state-action pair.
\ep 
 
 {\bf Simplification for null structure}. 
We conclude this section by proving that $\set{F}_{\TN{un}}(\phi) \subset \set{F}_\Phi(\phi)$. Let $\eta \in \set{F}_{\TN{un}}(\phi)$, recalling that

\begin{align*}
\set{F}_{\TN{un}}(\phi)
=
\left\{
\eta  \in\set{F}_0(\phi)
: 
 \eta(x, a) \left(\frac{\delta^*(x, a;\phi)}{ H+1}\right)^2 \ge 2, ~ 
\forall (x,a)~\text{s.t.}~a \notin \set{O}(x; \phi)
\right\} \;.
\end{align*}

We show that $\eta \in \set{F}_\Phi(\phi)$. To this aim, we need to show that $\forall (x,a)~\text{s.t.}~a \notin \set{O}(x; \phi),$

\begin{align*}
\eta(x, a) \kl_{\phi\mid \psi}(x,a) \ge 1, ~
\forall \psi \in \Delta_{\Phi}(x, a; \phi). 
\end{align*}

Let $(x,a)$ be such that $a\notin \set{O}(x;\phi)$, which means $a\notin \set{O}(x,h_\phi^\star;\phi)$, and $\psi  \in \Delta_\Phi(x, a ; \phi)$. We have, by definition, $(\mathbf{B}^{a}_\psi h^\star_\phi)(x) >
 g^\star_\phi + h^\star_\phi(x)$. Then,

\begin{align*}
\delta^*(x,a; \phi) &=  (\mathbf{B}^*_{\phi} h^*_\phi) (x)- (\mathbf{B}^a_{\phi} h^*_\phi) (x) \\
& <  (\mathbf{B}^a_{\psi} h^*_\phi) (x)- (\mathbf{B}^a_{\phi} h^*_\phi) (x) \\
&= r_\psi (x,a) - r_\phi (x,a) +
\sum_{y \in \set{S}} (p_\psi (y \mid x,a) - p_\phi (y \mid x,a)) h^*_\phi (y) \\
&\le  \| q_\psi (\cdot \mid x,a) -q_\phi (\cdot \mid x,a)\|_1
+ H \| p_\psi (\cdot \mid x,a) - p_\phi (\cdot \mid x,a)\|_1 \\
& \le (H+1) \|\psi(x,a) - \phi (x,a)\|_1
\end{align*}
where we define 
\begin{align*}
\|\psi(x,a) - \phi (x,a)\|_1 := \| q_\psi (\cdot \mid x,a) - q_\phi (\cdot \mid x,a)\|_1 + \| p_\psi (\cdot \mid x,a) - p_\phi (\cdot \mid x,a)\|_1.
\end{align*}
Finally, Pinsker's inequality yields: 
 \begin{align*}
2 \kl_{\phi \mid \psi}(x, a ) \ge \|\psi(x,a) - \phi(x,a)\|^2_1\ge  \left(\frac{\delta^*(x, a; \phi)}{H+1} \right)^2 .
\end{align*}
This implies that:
\begin{align*}
\eta(x,a)\kl_{\phi \mid \psi}(x, a ) \ge \frac{\eta(x,a)}{2}\left(\frac{\delta^*(x, a; \phi)}{H+1} \right)^2 \ge 1
\end{align*}
where the last inequality is due to the fact that $\eta(x,a) \in \set{F}_{\TN{un}}(\phi).$

\section{Proof of Theorem~\ref{thm:lip}} \label{sec:pf-lip}

We prove that $\set{F}_{\TN{lip}}(\phi) \subset \set{F}_{\Phi}(\phi)$. Let $\eta \in \set{F}_{\TN{lip}}(\phi)$. We show that $\eta \in \set{F}_{\Phi}(\phi)$. Let $\psi \in \Delta_\Phi(\phi)$, then, from \eqref{eq:opt2}, there exist $(x',a')$ such that $(\mathbf{B}^{a'}_\psi h^\star_\phi)(x') > g^\star_\phi + h^\star_\phi(x')$. Then, using the same arguments as at the end of the previous section, we obtain:

\begin{align} \label{eq:perturbed-prime}
\|\phi(x',a') - \psi (x',a')\|_1 \ge \frac{\delta^*(x', a'; \phi)}{H+1}.
\end{align}

Now for all $(x,a) \in \set{S} \times \set{A}$,

\begin{align}
\|\phi(x',a') - \psi (x',a')\|_1
& \le
\|\phi(x',a') - \phi (x,a)\|_1
+\|\phi(x,a) - \psi (x,a)\|_1
+\|\psi(x,a) - \psi (x',a')\|_1 \nonumber \\
&\le
\|\phi(x,a) - \psi (x,a)\|_1
+ 2L d(x, x')^\alpha
+ 2L' d(a, a')^{\alpha'}, \label{eq:use-lip}
\end{align}

where the first inequality follows from the triangular inequality and the second follows from Lipschitz continuity. This further implies that

\begin{align*}
\|\phi(x,a) - \psi (x,a)\|_1 \ge \left[ \frac{\delta^*(x', a'; \phi)}{H+1} - 2 \Big(L d(x, x')^\alpha
+ L' d(a, a')^{\alpha'}  \Big) \right]_+ .
\end{align*}
Hence, using Pinsker's inequality,
\begin{align} \label{eq:kl-lip-bound}
 2 \kl_{\phi \mid \psi}(x, a ) \ge \left[ \frac{\delta^*(x', a'; \phi)}{H+1} - 2 \Big(L d(x, x')^\alpha
+ L' d(a, a')^{\alpha'}  \Big) \right]_+^2,
\end{align}

which implies that:

\begin{align}
 \eta(x,a) \kl_{\phi \mid \psi}(x, a ) \ge \frac{\eta(x,a)}{2}\left[ \frac{\delta^*(x', a'; \phi)}{H+1} - 2 \Big(L d(x, x')^\alpha
+ L' d(a, a')^{\alpha'}  \Big) \right]_+^2 \ge 1.
\end{align}

The last inequality follows from $\eta \in \set{F}_{\TN{lip}}$. Thus $\set{F}_{\TN{lip}}(\phi) \subset \set{F}_{\Phi}(\phi)$.

\medskip
Next we derive an upper bound for $K_{\Phi}(\phi)$. To this aim, we construct a vector $\eta \ge 0$ verifying (2b) for our given structure $\Phi$. Then, we get an upper bound of $K_{\Phi}(\phi)$ by evaluating the objective function of $P(\phi,\set{F}_\Phi(\phi))$ at $\eta$.

To construct $\eta$, we build a sequence $(\set{X}_i)_{i=1,2,\ldots}$ of sets of (state, action) pairs, as well as a sequence $(x_i)_{i=1,2,\ldots}$(state, action) pairs, such that for any $i\ge 1$, $\set{X}_{i+1}\subset \set{X}_i$, and  $(x_i, a_i) \in \argmax_{(x,a) \in \set{X}_i} \delta^* (x,a; \phi)$ (ties are broken arbitrarily).

We start with $\set{X}_1 = \{(x,a) : x\in \set{S}, a \notin \set{O}(x; \phi), i.e., \delta^*(x, a;\phi) > 0 \}$. Recursively, for each $i =1, 2, ...$, let 
\begin{align}
\set{B}_i &= \left\{(x,a) \in \set{X}_i :  L d(x,x_i)^\alpha +L' d(a, a_i)^{\alpha'}  \le  \frac{\delta_{\min}}{4(H+1)} \right\} \;, \quad \text{and} \nonumber \\
\set{X}_{i+1} &= \set{X}_i \setminus \set{B}_i  \;. \label{eq:X-i}
\end{align}
Let $I$ be the first index such that $\set{X}_{I+1} = \emptyset$. Construct $\eta$ as
\begin{align} \label{eq:eta-construct}
\eta (x,a) = 
\begin{cases}
8  \left( \frac{\delta_{\min}}{H+1} \right)^{-2} &\text{if $\exists i \in [1, I]$ such that $(x,a) = (x_i, a_i)$},\\
0 &\text{otherwise.}
\end{cases} \;
\end{align}
Observe that $\eta$ is strictly positive at only $I$ pairs, and hence
\begin{align*}
\sum_{(x,a) \in \set{S} \times \set{A}}\delta^*(x,a ; \phi) \eta(x,a)  \le 
8  (H+1)  \left(\frac{H+1}{\delta_{\min}}\right)^2  I
\end{align*}
since $\delta^*(x,a; \phi) \le H+1 $ for all $(x,a)$. Next, we bound $I$ using the covering and packing numbers of
the hypercubes $[0, D]^d$ and $[0, D']^{d'}$.
\begin{lemma} \label{lem:I-bound}
The generation of $\set{X}_i$'s in \eqref{eq:X-i} 
must stop after $(S_{\textnormal{lip}}  A_{\textnormal{lip}}+1)$ iterations, i.e., $I \le S_{\textnormal{lip}}  A_{\textnormal{lip}}$.
\end{lemma}
The proof of this lemma is postponed at the end of this section. To complete the proof of the theorem, it remains to show that $\eta$ verifies all the constraints (2b) for the Lipschitz structure $\Phi$.

Remember that $\set{F}_{\TN{lip}}(\phi) \subset \set{F}_{\Phi}(\phi)$. Fix $\psi \in \Delta_\Phi (\phi)$. There exists $(x', a')$ such that $a' \notin \set{O}(x';h^*_\phi, \phi)$ and $a' \in \set{O}(x';h^*_\phi, \psi)$, and such that \eqref{eq:perturbed-prime} holds. Let $i \in \{1, \ldots, I\}$ denote an index such that $(x',a') \in \set{B}_i$.
Note that such an index $i$ exists since $(x',a') \in \set{X}_1$ and $\set{X}_{I+1} = \emptyset$.
Thus, we have:
\begin{align*}
\sum_{(x,a) \in \set{S} \times \set{A}} \eta(x, a) \kl_{\phi \mid \psi}(x,a) 
&\ge
\sum_{(x,a) \in \set{S} \times \set{A}}
\frac{\eta(x, a)}{2}\left[ \frac{\delta^*(x', a'; \phi)}{H+1} - 2 \Big(L d(x, x')^\alpha
+ L' d(a, a')^{\alpha'}  \Big) \right]_+^2 \\
&
\ge
\frac{\eta(x_i, a_i)}{2}\left[ \frac{\delta^*(x', a'; \phi)}{H+1} - 2 \Big(L d(x_i, x')^\alpha
+ L' d(a_i, a')^{\alpha'}  \Big) \right]_+^2  \\
&
\ge
\frac{\eta(x_i, a_i)}{2}\left[ \frac{\delta^*(x', a'; \phi)}{H+1} - 
\frac{1}{2} \frac{\delta_{\min}}{H+1} \right]_+^2 \\
&
\ge
4  \left( \frac{\delta_{\min}}{H+1} \right)^{-2} \left(\frac{1}{2} \frac{\delta_{\min}}{H+1} \right)^2 = 1
\end{align*}
where the third inequality follows from the fact that $(x',a') \in \set{B}_i$. Hence we have verified that $\eta$ satisfies the feasibility constraint for $\psi$. Since this observation holds for all $\psi \in\Delta_\Phi (\phi)$, this completes the proof of Theorem~\ref{thm:lip}.
\ep

{\bf Proof of Lemma~\ref{lem:I-bound}.} A $\delta$-{\em packing} of  a set $\set{D}$ with respect to a metric $\rho$ 
is a set $\{x_1, ..., x_n\} \subset \set{D}$ such that 
$\rho(x_i- x_j)  > \delta$ for all different $i, j \in \{1, ..., n\}$.
The {\em $\delta$-packing number} $I_{\textnormal{p}}(\delta, \set{D}, \rho)$ 
is the cardinality of the largest $\delta$-packing.
The construction of $\set{X}_i$ ensures that for different $i, j \in \{1, ..., I\}$,
\begin{align*}
\ell_{\TN{lip}}((x_i, a_i), (x_j, a_j) )
> \delta :=   \frac{\delta_{\min}}{4(H+1)}\;,
\end{align*}
where 
for $(x,a), (x',a') \in \mathbb{R}^d \times \mathbb{R}^{d'}$, 
\begin{align*}
\ell_{\TN{lip}}((x, a), (x', a') ) :=  L d(x, x')^\alpha +L' d(a, a')^{\alpha'}   \;.
\end{align*}
Then, we have:
\begin{align*}
I \le I_{\TN{p}} \left(\delta,
\set{S} \times \set{A},
  \ell_{\TN{lip}} \right )\;.
\end{align*}
To obtain an upper bound of the packing number, we further define
the covering number. A $\delta$-cover of a set $\set{D}$ with respect to a metric $\rho$ 
is a set $\{x_1, ..., x_I\} \subset \set{D}$ such that for each $x \in \set{D}$,
there exists some $i \in \{1, ..., I\}$ such that $\rho(x, x_i) \le \delta$.
The $\delta$-covering number $I_{\TN{c}}(\delta, \set{D}, \rho)$ is 
the smallest cardinality of $\delta$-cover. Then, we have the following relationship
between the packing and covering numbers.
\begin{lemma} \label{lem:cover-pack}
 For all $\delta >0$, $\set{D}, \set{D}'$ such that $\set{D} \subset \set{D}'$, 
 \begin{align*}
 I_{\TN{p}}(2 \delta, \set{D}, \rho) \le I_{\TN{c}}(\delta, \set{D}, \rho)
\le I_{\TN{c}}(\delta, \set{D}', \rho) \;.
\end{align*}
\end{lemma}
The proof of this lemma is provided at the end of the section for completeness. Define the metrics $\ell^{(1)}_{\max}, \ell^{(2)}_{\max}, \ell_{\max}$ for $\mathbb{R}^{d}, \mathbb{R}^{d'}, \mathbb{R}^{d} \times \mathbb{R}^{d'}$, respectively, as follows:
\begin{align*}
\ell^{(1)}_{\max}(x, x')
&:= \left( \frac{1}{\sqrt{d}} \left(\frac{\delta}{2L}\right)^{1/\alpha}  \right)^{-1} \| x - x' \|_{\infty}
\;, \\
\ell^{(2)}_{\max}(a, a') 
&:= \left( \frac{1}{\sqrt{d'}} \left(\frac{\delta}{2L'}\right)^{1/\alpha'}  \right)^{-1} \| a - a' \|_{\infty} 
\;, \\
\ell_{\max}((x,a), (x',a')) 
&:= \max \left\{  \ell^{(1)}_{\max}(x, x'),  \ell^{(2)}_{\max}(a, a') \right\} \;,
\end{align*}
where $\|\cdot\|_\infty$ is infinite norm.
Then, it follows that
for any $(x,a), (x',a')\in \mathbb{R}^d \times \mathbb{R}^{d'}$, 
\begin{align*}
\ell_{\max}((x,a), (x',a')) \le 1 \implies
\ell_{\TN{lip}}((x,a), (x',a')) \le  \delta \;.
\end{align*}
Hence, we have
 \begin{align*}
 I &\le I_{\TN{c}} (\delta, \set{S} \times \set{A}, \ell_{\TN{lip}})  \\
 &\le I_{\TN{c}} (1, \set{S} \times \set{A}, \ell_{\max})  \\ 
  &\le I_{\TN{c}} (1, \set{S}, \ell^{(1)}_{\max})  
  I_{\TN{c}} (1, \set{A}, \ell^{(2)}_{\max}) 
\end{align*}
since 
for any $1$-cover $\set{S}'$
of $\set{S}$ with metric $\ell^{(1)}_{\max}$ 
and 
any $1$-cover $\set{A}'$ of $\set{A}$ 
with metric $\ell^{(2)}_{\max}$,
their Cartesian product $\set{S}' \times \set{A}' = \{(x,a) : x \in \set{S}', a \in \set{A}' \}$
is $1$-cover of $\set{S} \times \set{A}$ with metric $\ell_{\max}$.
We now study $I_{\TN{c}} (1, \set{S}, \ell^{(1)}_{\max})$ and $I_{\TN{c}} (1, \set{A}, \ell^{(2)}_{\max})$.
Recalling $\set{S} \subset [0,D]^d$ and using Lemma~\ref{lem:cover-pack}, it follows directly that
\begin{align*}
I_{\TN{c}} (1, \set{S}, \ell^{(1)}_{\max}) &\le
I_{\TN{c}} (1, [0, D]^d, \ell^{(1)}_{\max}) \\
& = I_{\TN{c}} \left( \frac{1}{\sqrt{d}} \left( \frac{\delta}{2L} \right)^{1/\alpha}, [0, D]^d, \|\cdot \|_{\infty} \right)  \\
&\le \left( \frac{D}{\frac{1}{\sqrt{d}} \left( \frac{\delta}{2L} \right)^{1/\alpha}} +1 \right)^d\;,
\end{align*}
which implies
\begin{align*}
I_{\TN{c}} (1, \set{S}, \ell^{(1)}_{\max}) 
&\le  
\min \left\{|\set{S}|, 
\left( \frac{D}{\frac{1}{\sqrt{d}} \left( \frac{\delta}{2L} \right)^{1/\alpha}} +1 \right)^d
\right\}  = S_{\TN{lip}}
\end{align*}
where we used the fact that $I_{\TN{c}} (1, \set{S}, \ell^{(1)}_{\max}) \le |\set{S}|$.
Similarly, we have
\begin{align*}
I_{\TN{c}} (1, \set{A}, \ell^{(2)}_{\max}) 
&\le  
\min \left\{|\set{A}|, 
\left( \frac{D'}{\frac{1}{\sqrt{d'}} \left( \frac{\delta}{2L'} \right)^{1/\alpha'}} +1 \right)^{d'}
\right\}  = A_{\TN{lip}}\;.
\end{align*}
This completes the proof of Lemma~\ref{lem:I-bound}.\ep

\medskip
{\bf Proof of Lemma~\ref{lem:cover-pack}.} Consider a $\delta$-cover $\set{X}$ and a $2\delta$-packing $\set{Y}$
of set $\set{D}$ with respect to metric $\rho$.
Then, there is no $x \in \set{X}$ such that $y, y' \in \set{B}(\delta, x) = \{x' \in \set{D}: \rho(x, x') \le \delta \}$
for two different $y, y' \in \set{Y}$. Otherwise, we would have
$\rho(x, y) \le \delta$ and
$\rho(x, y') \le \delta$ which
implies $\rho(y, y') \le 2\delta$ from the triangle inequality, and contradicts the fact that $y, y'$ are two different elements of $2\delta$-cover, i.e., $\rho(y, y') > 2\delta$. 
Thus, the cardinality of $\set{Y}$ cannot be larger than that of $\set{X}$. Due to the arbitrary choice of 
$\delta$-cover $\set{X}$ and a $2\delta$-packing $\set{Y}$, we conclude that $I_{\TN{p}}(2 \delta, \set{D}, \rho) \le I_{\TN{c}}(\delta, \set{D}, \rho)$.

The second inequality in the lemma is straightforward.\ep

\section{Proof of Theorem~\ref{thm:DEL}}

We analyze the regret under $\pi=$ DEL algorithm when implemented with the original feasible set $\set{F}_\Phi(\phi; {\cal C},\zeta)$. Extending the analysis to the case where DEL runs on the simplified feasible sets $\set{F}_{\textnormal{un}}(\phi; {\cal C},\zeta)$ and $\set{F}_{\textnormal{lip}}(\phi; {\cal C},\zeta)$ can be easily done.

For $T \ge 1$, $\varepsilon > 0$, $x \in \set{S}$ and $a \in \set{A}$, define the following random variables:
\begin{align*}
W^{(1)}_T (x,a; \varepsilon) 
&:= \sum_{t = 1}^T 
\mathbbm{1} \left[
(X_t, A_t) = (x,a),  \set{E}_t(\varepsilon),
(\mathbf{B}^a_{\phi_t} h'_t)(x) \le  (\mathbf{B}^*_{\phi} h^*_\phi)(x) - 2\varepsilon
\right] \\
W^{(2)}_T (x,a; \varepsilon) 
&:= \sum_{t = 1}^T 
\mathbbm{1} \left[
(X_t, A_t) = (x,a),  \set{E}_t(\varepsilon), 
(\mathbf{B}^a_{\phi_t} h'_t)(x) >  (\mathbf{B}^*_{\phi} h^*_\phi)(x) - 2\varepsilon
\right] \\
W^{(3)}_T (\varepsilon) 
&:= \sum_{t = 1}^T 
\mathbbm{1} \left[
\neg\set{E}_t(\varepsilon)
\right]
\end{align*}
where we use the standard notation $\neg\set{U}$ to represent the event that $\set{U}$ does not occur, where we recall that $h'_t:=h_{\phi_t'}$ is the bias function of the restricted estimated model $\phi_t'=\phi_t({\cal C}_t)$ at time $t$, and where the event $\set{E}_t(\varepsilon)$ is defined as:
\begin{align*}
\set{E}_t(\varepsilon) :=
\left\{ 
\Pi^*(\phi'_t) \subseteq \Pi^*(\phi)
~\text{and}~
|r_t(x,a) - r_{\phi}(x,a) | + |h'_t (x) - h^*_\phi(x)| \le \varepsilon 
~\forall x \in \set{S}, \forall a \in \set{O}(x;\phi'_t)
\right\} \;.
\end{align*}
From the above definitions, we have:
\begin{subequations} \label{eq:R-decomp}
\begin{align}
R^\pi_T( x_1) &\le   
\sum_{(x, a): a \notin \set{O}(x;\phi)}
\delta^*(x,a; \phi) \EXP^\pi_{\phi \mid x_1}\left[W^{(1)}_T(x,a; \varepsilon) \right]   \label{eq:R-decomp-1}\\
&\quad + 
\sum_{(x, a): a \notin \set{O}(x;\phi)}
S \EXP^\pi_{\phi \mid x_1}\left[W^{(2)}_T(x,a; \varepsilon) \right]  \label{eq:R-decomp-2}\\
&\quad + S\EXP^\pi_{\phi \mid x_1} [W^{(3)}_T(\varepsilon)] \;.\label{eq:R-decomp-3}
\end{align}
\end{subequations}
The multiplicative factor $S$ in the last two terms arises from the fact that $\max_{(x,a)} \delta^*(x,a; \phi) \le S$
when the magnitude of the instantaneous reward is bounded by $1$.
%
%
%
%
%
%
Next we provide upper bounds of each of the three terms in \eqref{eq:R-decomp}.

{\bf A. Upper bounds for (\ref{eq:R-decomp-1}) and (\ref{eq:R-decomp-2}).}
To study the first two terms in \eqref{eq:R-decomp}, we first make the following observations on the behavior of the algorithm. Let $\set{E}^\est_t$, $\set{E}^\mnt_t$, $\set{E}^\xpt_t$, and $\set{E}^\xpr_t$ denote the events that at time $t$, the algorithm enters the estimation, monotonization, exploitation, and exploration phases, respectively.
 By the design of the algorithm, the estimation phase generates regret no more than $O(\log T / \log \log T) = o(\log T)$, i.e.,
\begin{align}
\sum_{t=1}^T \mathbbm{1}[ \set{E}^{\est}_t] = o(\log T).  \label{eq:sublog-est}
\end{align}
Moreover, when the event $\set{E}_t(\varepsilon)$ occurs, we have $\set{O}(X_t ; \phi'_t) 
\subseteq \set{O}(X_t; \phi)$ and thus the monotonization and exploitation phases produce no regret. Formally, for $(x,a)  \in \set{S} \times \set{A}$ such that $a \notin \set{O}(x; \phi)$,
\begin{align}
\sum_{t=1}^T \mathbbm{1}[ 
(X_t, A_t) = (x,a), \set{E}_{t}(\varepsilon), \set{E}^{\mnt}_t \cup \set{E}^{\xpt}_t
] = 0 . \label{eq:on-E}
\end{align}

Hence, when $\set{E}_t(\varepsilon)$ occurs, we just care about the regret generated in the exploration phase, i.e., 
for any $(x,a) \in \set{S} \times \set{A}$ such that $a \notin \set{O}(x; \phi)$,
\begin{align*}
 \EXP^\pi_{\phi \mid x_1}\left[W^{(1)}_T(x,a; \varepsilon) \right] 
 &\le o(\log T)+  \sum_{t=1}^T \Pr^\pi_{\phi \mid x_1}\left[ \set{Z}^{(1)}_t(x,a; \varepsilon) \right] \\
\EXP^\pi_{\phi \mid x_1}\left[W^{(2)}_T(x,a; \varepsilon) \right] &\le o(\log T) +\sum_{t=1}^T \Pr^\pi_{\phi \mid x_1}\left[ \set{Z}^{(2)}_t(x,a; \varepsilon) \right],
\end{align*}
where  the events $\set{Z}^{(1)}_t (x,a; \varepsilon)$ and $\set{Z}^{(2)}_t (x,a; \varepsilon)$ are defined as:
\begin{align*}
\set{Z}^{(1)}_t (x,a; \varepsilon) &:= 
 \left\{ (X_t, A_t) = (x,a), \set{E}_{t}(\varepsilon), \set{E}^{\xpr}_t,
 (\mathbf{B}^a_{\phi_t} h'_t)(x) \le  (\mathbf{B}^*_{\phi} h^*_\phi)(x) - 2\varepsilon
  \right\}  \\
 \set{Z}^{(2)}_t (x,a; \varepsilon) &:= 
 \left\{ (X_t, A_t) = (x,a), \set{E}_{t}(\varepsilon), \set{E}^{\xpr}_t,
 (\mathbf{B}^a_{\phi_t} h'_t)(x) >  (\mathbf{B}^*_{\phi} h^*_\phi)(x) - 2\varepsilon
  \right\}. 
\end{align*}


The following lemma is proved in Section~\ref{sec:pf-bound-1}, and deals events $\set{Z}^{(1)}_t (x,a; \varepsilon)$.

\medskip

\begin{lemma}
\label{lem:bound-1}
For structure $\Phi$ with Bernoulli rewards and an ergodic MDP $\phi \in \Phi$,
consider $\pi = \textnormal{DEL($\gamma$)}$ for $\gamma > 0$.
Suppose that ($i$) $\phi$ is in the interior of $\Phi$; 
($ii$) the solution $\eta^*(\phi)$ is unique for each $(x,a)$ such that $a  \notin \set{O} (x;\phi)$; 
and ($iii$) continuous at $\phi$.
Then, for any $(x,a) \in \set{S} \times \set{A}$ such that $a \notin \set{O}(x; \phi)$, 
\begin{align*}
\lim_{\varepsilon \to 0} \limsup_{T \to \infty}
\frac{ \sum_{t =1}^T \Pr^\pi_{\phi \mid x_1}\left[ \set{Z}^{(1)}_t(x,a; \varepsilon) \right]}{\log T} \le (1+\gamma) \eta^*(x,a;\phi) \;.
\end{align*}
\end{lemma}

The following lemma is proved in Section~\ref{sec:pf-bound-2}, and deals events $\set{Z}^{(2)}_t (x,a; \varepsilon)$. Its proof relies on the following observation. When $\set{Z}^{(2)}_t(x,a;\varepsilon)$ occurs for sufficiently small $\varepsilon$, the facts that $\set{E}_t(\varepsilon)$ holds and that $ (\mathbf{B}^a_{\phi_t} h'_t)(x) < (\mathbf{B}^*_{\phi} h^*_\phi)(x) - 2\varepsilon$ imply that $\phi_t(x,a)$ does not estimate $\phi(x,a)$ accurately. The lemma then follows from concentration arguments.

\medskip

\begin{lemma}
\label{lem:bound-2}
For structure $\Phi$ with Bernoulli rewards and an ergodic MDP $\phi \in \Phi$,
consider $\pi = \textnormal{DEL($\gamma$)}$ for $\gamma > 0$.
Then, there exists  $\varepsilon_2 >0 $ such that
for any $(x,a) \in \set{S} \times \set{A}$ such that $a \notin \set{O}(x; \phi)$
and $\varepsilon \in (0, \varepsilon_2)$, 
\begin{align*}
\sum_{t =1}^T \Pr^\pi_{\phi \mid x_1}\left[ \set{Z}^{(2)}_t(x,a; \varepsilon) \right] = o(\log T)
\quad \text{ as  $T \to \infty$.}
\end{align*}
\end{lemma}

{\bf B. Upper bound for (\ref{eq:R-decomp-3}).} The last term in \eqref{eq:R-decomp} is concerned with the regret generated when $\set{E}_t(\varepsilon)$ does not occur. It is upper bounded in the following lemma proved in Section~\ref{sec:pf-bound-3}. To establish this result, we use a similar argument as that in Proposition~5 of \cite{burnetas1997optimal}. Intuitively, we show that by the design of the algorithm, the restricted bias function $h'_t$ is monotonically improved so that it eventually converges to the optimal bias function $h^*_\phi$ with high probability. In this analysis, we provide a more sophisticated concentration inequality than the one in \cite{burnetas1997optimal}. This concentration inequality is particularly important to bound the regret generated in the exploitation phase. 

\begin{lemma} \label{lem:bound-3}
For structure $\Phi$ with Bernoulli rewards and an ergodic MDP $\phi \in \Phi$,
consider $\pi = \textnormal{DEL($\gamma$)}$ for $\gamma > 0$.
Suppose $\phi$ is in the interior of $\Phi$, i.e., 
there exists a constant $\zeta_0 >0$ such that for any $\zeta \in (0, \zeta_0)$,
$\psi \in \Phi$ if $\|\phi - \psi\| \le \zeta$.
Then, there exists $\varepsilon_3 > 0$ such that for any $\varepsilon \in (0, \varepsilon_3)$, 
\begin{align}
\Pr^{\pi}_{\phi \mid x_1 } [ \neg\set{E}_T(\varepsilon)]  =  o(1/T) 
\quad \text{as $T \to \infty$}.
\label{eq:wts-W-3}
\end{align}
\end{lemma} 
We provide the proof of Lemma~\ref{lem:bound-3} in Section~\ref{sec:pf-bound-3}.
Now, we are ready to complete the proof of Theorem~\ref{thm:DEL}. 
Combining Lemma~\ref{lem:bound-1}, \eqref{eq:on-E} and \eqref{eq:sublog-est}, we get
\begin{align*}
\sum_{x\in \set{S}} \sum_{a \notin \set{O}(x;\phi)}
\delta^*(x,a;\phi)
\left(
\lim_{\varepsilon \to 0} \limsup_{T \to \infty}
\frac{\EXP^\pi_{\phi \mid x_1}\left[ W^{(1)}_T(x,a; \varepsilon) \right]}{\log T} 
\right)
&\le (1+\gamma)\sum_{x\in \set{S}} \sum_{a \in \set{A}} \delta^*(x,a;\phi) \eta^*(x,a;\phi)  \nonumber \\
&= (1+\gamma) K_{\Phi}(\phi)  .
\end{align*}
Similarly, combining Lemma~\ref{lem:bound-2} with \eqref{eq:on-E} and \eqref{eq:sublog-est},
it follows that for sufficiently small $\varepsilon \in (0, \min \{\varepsilon_2, \varepsilon_3\})$, 
\begin{align*}
\limsup_{T \to \infty}
\frac
{ \EXP^\pi_{\phi \mid x_1}\left[
\sum_{x\in \set{S}} \sum_{a \notin \set{O}(x;\phi)} 
S W^{(2)}_T(x,a; \varepsilon) \right]}
{\log T} = 0 .
\end{align*}
From Lemma~\ref{lem:bound-3}, we have that for sufficiently small $\varepsilon \in (0, \min \{\varepsilon_2, \varepsilon_3\})$, 
\begin{align*}
\limsup_{T \to \infty}
\frac
{ \EXP^\pi_{\phi \mid x_1}\left[
 W^{(3)}_T (\varepsilon) \right]}
{\log T} = 0 . 
\end{align*}
Therefore, recalling the decomposition of regret bound in \eqref{eq:R-decomp}, we conclude the proof of Theorem~\ref{thm:DEL}.

\ep


\subsection{Proof of Lemma~\ref{lem:bound-1}}
\label{sec:pf-bound-1}


To establish the lemma, we investigate the event $\set{Z}^{(1)}_t(x,a; \varepsilon)$ depending on whether $\set{F}_t$ is empty or not, and on whether $\phi_t$ is a good approximation of $\phi$. To this aim, for any given $t >0$ and $\zeta >0$, define the event
$ \set{B}_t (\zeta) :=  \bigcap_{(x,a) \in \set{S} \times \set{A}} \set{B}_t(x,a; \zeta)$
where for each $(x,a) \in \set{S} \times \set{A}$, $\set{B}_t (x,a ; \zeta) := \{\|  \phi_t(x,a) - \phi(x,a) \| \le \zeta \}$. 
Fix $(x,a) \in \set{S} \times \set{A}$ such that $a \notin \set{O}(x ; \phi)$. By the continuity assumption made in Theorem \ref{thm:DEL}, we have:
\begin{align*}
&
\sum_{t=1}^T   
\Pr^\pi_{\phi | x_1} \left[  
\set{Z}^{(1)}_t(x,a; \varepsilon), 
\set{F}_t \neq \emptyset, \zeta_t < \zeta(\varepsilon), \set{B}_t(\zeta_t) 
 \right] 
 \\
 &\le 
 \EXP^\pi_{\phi | x_1} \left[  
\sum_{t=1}^T   
\mathbbm{1} \left[
(X_t, A_t) = (x,a),
N_t(x,a) \le \eta_t(x,a) \gamma_t, 
\set{F}_t \neq \emptyset,
\zeta_t < \zeta(\varepsilon),
 \set{B}_t(\zeta_t)
 \right]
 \right]
 \\
 &\le 
 \EXP^\pi_{\phi | x_1} \left[  
\sum_{t=1}^T   
\mathbbm{1} \left[
(X_t, A_t) = (x,a),
N_t(x,a) \le (\eta^*(x,a ;\phi)+\varepsilon) \gamma_t
 \right]
 \right]
  \\
 &\le 
(\eta^*(x,a ;\phi)+\varepsilon) \gamma_t + 2
\end{align*}
where the second inequality is from the continuity of $\eta^*(\phi)$,
and the last inequality is from a simple counting argument made precise in the following lemma  \cite{burnetas1997optimal} (Lemma~3 therein):
\begin{lemma} \label{lem:counting}
Consider any (random) sequence of $Z_t \in \{0, 1\}$ for $t >0$. Let $N_T := \sum_{t=1}^T \mathbbm{1}[Z_t = 1]$. Then, for all $N >0$, 
$ \sum_{t=1}^T \mathbbm{1} [Z_t = 1, N_t \le N] \le N +1$
(point-wise if the sequence is random).
\end{lemma}
{\bf Proof of Lemma~\ref{lem:counting}.}
The proof is straightforward from rewriting the summation as follows:
\begin{align*}
\sum_{t=1}^T \mathbbm{1} [Z_t = 1, N_t \le N]
&= \sum_{t=1}^T \sum_{n=1}^{\lfloor N \rfloor}\mathbbm{1} [Z_t = 1, N_t = n] \\
& =  \sum_{n=1}^{\lfloor N \rfloor}   \sum_{t=1}^T \mathbbm{1} [Z_t = 1, N_t = n]
\le  N  +1
\end{align*}
where the last inequality is from the fact that $ \sum_{t=1}^T \mathbbm{1} [Z_t = 1, N_t = n] \le 1$.
\ep 

Since $\lim_{T \to \infty} \frac{\gamma_T}{\log T} = (1+\gamma)$ for all $x \in \set{S}$, we obtain:
\begin{align*}
\lim_{\varepsilon \to 0} \limsup_{T \to \infty}
\frac
{
\EXP^\pi_{\phi | x_1} \left[  
\sum_{t=1}^T   
\mathbbm{1} \left[
\set{Z}^{(1)}_t(x,a; \varepsilon),
\set{F}_t \neq \emptyset, \zeta_t < \zeta(\varepsilon), \set{B}_t(\zeta_t) 
 \right]
 \right] 
 }
 {\log T}
 = (1+\gamma)\eta^*(x,a;\phi) .
\end{align*}
Hence, to complete the proof of Lemma ~\ref{lem:bound-1}, it suffices to show that
\begin{align}
\sum_{t=1}^T   
\Pr^\pi_{\phi | x_1} \left[  
\set{Z}^{(1)}_t(x,a; \varepsilon),
\set{F}_t = \emptyset \right]  &= O(1)  \label{eq:bound-1-wts-1}\\
 \sum_{t=1}^T   
\Pr^\pi_{\phi | x_1} \left[  
\set{Z}^{(1)}_t(x,a; \varepsilon),
\set{F}_t \neq \emptyset, 
\neg\set{B}_t(\zeta_t)
 \right]  &= o(\log T)  \label{eq:bound-1-wts-2}
\end{align}
since $\sum_{t=1}^T   
\Pr^\pi_{\phi | x_1} \left[  
\zeta_t >\zeta(\varepsilon)
 \right]  = O(1)$.

To prove \eqref{eq:bound-1-wts-1},
observe that on the event $\set{Z}^{(1)}_t (x,a; \varepsilon)$
for sufficiently large $t \ge e^{e^\varepsilon}$, i.e., $\zeta_t < \varepsilon$, 
for $b \in \set{O}(x; \phi'_t)$, we have
\begin{align*}
\delta^*(x,a; \phi_t, \set{C}_t) &= 
(\mathbf{B}^b_{\phi'_t} h'_t)(x)- (\mathbf{B}^a_{\phi_t} h'_t)(x) \\
&\ge (\mathbf{B}^b_{\phi'_t} h'_t)(x) -  (\mathbf{B}^*_{\phi} h^*_\phi)(x) + 2\varepsilon
\\
&\ge -|(\mathbf{B}^b_{\phi'_t} h'_t)(x) -  (\mathbf{B}^b_{\phi} h^*_\phi)(x)| + 2\varepsilon
\\
&\ge - (|r_t(x, b) - r_\phi (x,b)| + |h'_t(x) - h^*_\phi (x)|) + 2\varepsilon
\\
 &\ge \varepsilon > \zeta_t
\end{align*}
where the first, second, and fourth inequalities are from 
that on the event $\set{Z}^{(1)}_t (x,a; \varepsilon)$,
$(\mathbf{B}^a_{\phi_t} h'_t)(x) \le  (\mathbf{B}^*_{\phi} h^*_\phi)(x) - 2\varepsilon$,
$\set{O}(x; \phi'_t) \subseteq \set{O}(x; \phi)$, and
$|r_t(x, b) - r_\phi (x,b)| + |h'_t(x) - h^*_\phi (x)| \le \varepsilon$, respectively,
and the last one is from the choice of $t$ such that $\zeta_t = {1}/{(1+\log \log t)} < \varepsilon$.
Therefore, when $\set{Z}^{(1)}_t (x,a; \varepsilon)$ occurs for sufficiently large $t \ge e^{e^\varepsilon}$,
\begin{align} \label{eq:delta_t-bound}
\delta_t(x,a) >\zeta_t  > 0 .
\end{align}
If $\set{F}_t$ is empty, from the design of DEL algorithm, $\delta_t(x,a) > 0$ implies that $\eta_t(x,a) = 0$ and thus $(x,a)$ is not selected in the exploration phase. This concludes the proof of \eqref{eq:bound-1-wts-1} as  $\sum_{t=1}^T   
\Pr^\pi_{\phi | x_1} \left[  \zeta_t >\varepsilon  \right]  \le  e^{e^\varepsilon} =  O(1)$.

To show \eqref{eq:bound-1-wts-2}, observe that when $\set{Z}^{(1)}_t (x,a; \varepsilon)$ and $\set{F}_t\neq\emptyset$ occur, for $t \ge e^{e^\varepsilon}$ combining \eqref{eq:delta_t-bound} and Lemma~\ref{lem:sol-bound} given below, we get:
\begin{align}
\eta_t(x,a) \le 2SA \left(\frac{S+1}{\zeta_t}\right)^2 . \label{eq:sol-bound}
\end{align}
 \begin{lemma} \label{lem:sol-bound}
 Consider a structure $\Phi$, an MDP $\phi \in \Phi$, a non-empty correspondence $\set{C} : \set{S} \twoheadrightarrow \set{A}$, and $\zeta >0$. If $\set{F}_\Phi(\phi;  \set{C}, \zeta) $ is non-empty
 and there exists $(x, a) \in \set{S} \times  \set{A}$ such that $\delta^*(x, a ; \phi, \set{C}, \zeta)  >0$,
 then $\eta^*(x,a; \phi, \set{C}, \zeta) \le  2SA \left(\frac{S+1}{\zeta}\right)^2$ where
 $\eta^*(x,a; \phi, \set{C}, \zeta) $ is a solution of $\set{P}(\delta^*(\phi, \set{C}, \zeta),\set{F}_\Phi(\phi;  \set{C}, \zeta))$.
\end{lemma}
{\bf Proof of Lemma~\ref{lem:sol-bound}. } 
Using the same arguments as those used in Theorem~\ref{thm:unrelated}
to show that $\set{F}_{\textnormal{un}} (\phi) \subset  \set{F}_{\Phi} (\phi)$, 
one can easily check that $\set{F}_{\textnormal{un}} (\phi ; \set{C}, \zeta) \subset  \set{F}_{\Phi} (\phi ; \set{C}, \zeta)$. Note that the diameter of bias function with Bernoulli reward is bounded by $S$. Now for $(x,a) \in \set{S} \times \set{A}$ such that $\delta^*(x,a; \phi, \set{C}, \zeta)>0$, we have 
\begin{align} \label{eq:detla-zeta-bound}
\delta^*(x,a; \phi, \set{C}, \zeta) > \zeta
\end{align}
which then implies that $2 \left( \frac{H_{\phi(\set{C})}+1}{\delta^*(x,a; \phi, \set{C}, \zeta)}\right)^2  \le 
2 \left( \frac{S+1}{\zeta}\right)^2 $.
Now let $\eta$ be defined as $\eta(x,a) = \infty$ if $\delta^*(x,a; \phi, \set{C}, \zeta) =0$ and $\eta(x,a) = 2 \left( \frac{S+1}{\zeta}\right)^2$ otherwise. Then $\eta\in \set{F}_{\textnormal{un}} (\phi ; \set{C}, \zeta) \subset  \set{F}_{\Phi} (\phi ; \set{C}, \zeta)$. We deduce that the optimal objective value of $\set{P}(\delta^*(\phi, \set{C}, \zeta),\set{F}_\Phi(\phi;  \set{C}, \zeta))$ is upper-bounded by 
\begin{align*}
\sum_{(x,a) \in \set{S} \times \set{A}}  \eta^*(x,a; \phi, \set{C}, \zeta) \delta^*(x,a; \phi, \set{C}, \zeta) 
&\le \sum_{(x,a) \in \set{S} \times \set{A}}  \eta(x,a) \delta^*(x,a; \phi, \set{C}, \zeta) \\
&\le 2SA \frac{(S+1)^2}{\zeta} . 
\end{align*}
Using the optimality of $\eta^*(\phi, \set{C}, \zeta)$
 and \eqref{eq:detla-zeta-bound}, 
we conclude that 
for $(x,a) \in \set{S} \times \set{A}$ such that $\delta^*(x,a; \phi, \set{C}, \zeta)>0$, 
$ \eta^*(x,a; \phi, \set{C}, \zeta)  \le 2SA \left( \frac{S+1}{\zeta} \right)^2$. 
\ep 

From \eqref{eq:sol-bound}, we deduce by design of DEL that, if $\set{Z}^{(1)}_t (x,a; \varepsilon)$ and $\set{F}_t\neq\emptyset$ occur, for $t \ge e^{e^\varepsilon}$, then:
\begin{align*}
N_t(x,a) &\le \eta_t(x,a)\gamma_t\\
&\le 2SA \left(\frac{S+1}{\zeta_t}\right)^2 \gamma_t \le \gamma_t'
\end{align*}
 where
 \begin{align} \label{eq:def-gamma_t-prime}
 \gamma'_t  := 8S^3 A (1+\gamma) (1+\log\log t)^2 (\log t + 1)
 > 2SA \left(\frac{S+1}{\zeta_t} \right)^2 \gamma_t.
\end{align}
Hence defining $\set{B}'_t(x,a) := \left\{  (X_t,A_t) = (x,a), N_t(x,a) \le \gamma'_t, 
 \neg\set{B}_t(\zeta_t)  \right\}$, we get:
\begin{align*}
 \sum_{t=1}^T   
\Pr^\pi_{\phi | x_1} \left[  
\set{Z}^{(1)}_t(x,a; \varepsilon),
\set{F}_t \neq \emptyset, 
\neg\set{B}_t(\zeta_t)
 \right]  
&\le 
  \sum_{t=1}^T   
\Pr^\pi_{\phi | x_1} \left[  
\set{Z}^{(1)}_t(x,a; \varepsilon),
\set{F}_t \neq \emptyset, 
\neg\set{B}_t(\zeta_t),
t \ge e^{e^\varepsilon}
 \right]  
 + e^{e^\varepsilon}
\\
&\le 
  \sum_{t=1}^T   
\Pr^\pi_{\phi | x_1} \left[ \set{B}'_t(x,a) \right] 
+ O(1) .
 \end{align*}

Using  $\rho >0$ in \eqref{eq:erg}, we check that 
\begin{align}
& 
\sum_{t=1}^T   
\Pr^\pi_{\phi | x_1} \left[  
\set{B}'_t(x,a) 
 \right] \nonumber \\
 &\le 
 \sum_{t=1}^T   
\Pr^\pi_{\phi | x_1} \left[  
\min_{y\in \set{S}}N_t(y) \ge \rho t,
\set{B}'_t(x,a) 
 \right]  
 +
  \sum_{t=1}^T   
\Pr^\pi_{\phi | x_1} \left[  
\min_{y\in \set{S}}N_t(y) \le \rho t 
\right]   \nonumber\\
 &\le 
 \sum_{t=1}^T   
\Pr^\pi_{\phi | x_1} \left[  
\min_{y\in \set{S}}N_t(y) \ge \rho t,
\set{B}'_t(x,a) 
 \right]  
 + o (\log T)  \nonumber \\
 &\le \sum_{t=1}^T   
\Pr^\pi_{\phi | x_1} \left[  
\min_{(y,b)\in \set{S}\times \set{A}}N_t(y,b) \ge \frac{\log  t}{(1+\log \log t)^2}, 
\set{B}'_t(x,a) 
 \right]  + o(\log T). 
 \label{eq:erg-app-Z1} 
\end{align}
Here, the second inequality stems from \eqref{eq:erg} and a union bound (over states). The the last inequality follows from the following lemma: 
\begin{lemma} \label{lem:mono-est-Z1}
Under DEL algorithm, we have 
\begin{align}
& \sum_{t=1}^T   
\mathbbm{1} \left[  
\min_{y\in \set{S}}N_t(y) \ge \rho t,
\min_{(y,b)\in \set{S}\times \set{A}}N_t(y,b) < \frac{\log  t}{(1+\log \log t)^2}
 \right]   =  o(\log T) \label{eq:mono-est-Z1} .
\end{align}
\end{lemma}
{\bf Proof of Lemma~\ref{lem:mono-est-Z1}.}
For $(x,a) \in \set{S} \times \set{A}$ and $t$ sufficiently large, we claim the following:
\begin{align}
\mathbbm{1} \left[  
N_t(x) \ge \rho t,
N_t(x,a) < \frac{\log  t}{(1+\log \log t)^2}
 \right]   =  0 .\label{eq:claim}
\end{align}
Using the above claim, we can complete the proof. Indeed:
\begin{align*}
& \sum_{t=1}^T   
\mathbbm{1} \left[  
\min_{y\in \set{S}}N_t(y) \ge \rho t,
\min_{(y,b)\in \set{S}\times \set{A}}N_t(y,b) < \frac{\log  t}{(1+\log \log t)^2}
 \right] \\
 & \le  \sum_{t=1}^T   
 \sum_{(x,a) \in \set{S} \times \set{A}}
\mathbbm{1} \left[  
\min_{y\in \set{S}}N_t(y) \ge \rho t,
N_t(x,a) < \frac{\log  t}{(1+\log \log t)^2}
 \right] \\
  & \le  \sum_{t=1}^T   
 \sum_{(x,a) \in \set{S} \times \set{A}}
\mathbbm{1} \left[  
N_t(x) \ge \rho t,
N_t(x,a) < \frac{\log  t}{(1+\log \log t)^2}
 \right] =  O(1)  \quad \text{as $T \to \infty$.}
\end{align*} 
where the first inequality stems from the union bound.

Next we prove the claim \eqref{eq:claim}. Fix $(x,a) \in \set{S} \times \set{A}$ and consider sufficiently large $t$.
Suppose $N_t(x) \ge \rho t$ and let $t_0 = \lfloor \rho t/2 \rfloor$. Then, since $N_{t_0}(x) \le t_0$,
it follows that
\begin{align*}
N_t(x) - N_{t_0} (x) \ge \rho t - \lfloor \rho t/ 2 \rfloor \ge \rho t/2 .
\end{align*}
Let $t_1  = \min \{u \in \mathbb{N} : u \in [t_0, t], N_{u}(x)-N_{t_0}(x) = \lfloor \rho t/4 \rfloor\}$ denote the time when the number of visits to state $x$ after time $t_0$ reaches $\lfloor \rho t/4 \rfloor$.
Since $N_t(x) - N_{t_0} (x) \ge \rho t/2 \ge \lfloor \rho t/4 \rfloor$, there exists such a $t_1 \in [t_0, t]$.
From the construction of $t_1$, it follows that for all $u \in [t_1, t]$, $\lfloor \rho t/4 \rfloor \le  N_u (x)$ and
\begin{align}
N_t(x) - N_{t_1}(x) &= \left(N_t(x) - N_{t_0}(x) \right) - \left(N_{t_1}(x) - N_{t_0}(x) \right) \nonumber \\
&\ge \rho t /2 - \lfloor \rho t / 4 \rfloor \ge \rho t  / 4 \label{eq:N-t-N-t1}.
\end{align}

Let $\set{N}_{t_1, t}(x) := \{u \in [t_1, t] : X_u = x, \neg\set{E}_u^\mnt \}$ be the set of times between $t_1$ and $t$ when the state is $x$ and the algorithm does not enter the monotonization phase and hence checks the condition to enter the estimation phase. For $u \in\set{N}_{t_1, t}(x)$, the condition for the algorithm to enter the estimation phase and select an action with the minimum occurrence is: 
\begin{align} \label{eq:must-enter-est}
\exists b \in \set{A}:\ N_u(x, b) <  
\frac{\log \lfloor \rho t/4 \rfloor}{1 + \log \log \lfloor \rho t/4 \rfloor} 
\end{align}
since from the construction of $t_1$, for any $u \in [t_1, t]$, we have $\frac{\log \lfloor \rho t/4 \rfloor}{1 + \log \log \lfloor \rho t/4 \rfloor} 
\le
\frac{\log N_u(x)}{1 + \log \log N_u(x)}$.

Now assume that the number of times the algorithm enters the monotonization phase in state $x$ between $t_1$ and $t$ is bounded by $O(\log t)$. From \eqref{eq:N-t-N-t1} and \eqref{eq:must-enter-est}, we deduce the desired claim \eqref{eq:claim}. Indeed, with the observation \eqref{eq:must-enter-est}, the fact that monotonization happens a sublinear number of times implies that the algorithm estimates all actions more than $\frac{\log \lfloor \rho t/4 \rfloor}{1 + \log \log \lfloor \rho t/4 \rfloor}$ ($> \frac{\log t}{(1+\log \log t)^2}$) times. Actually, the fact that monotonization happens a sublinear number of times and \eqref{eq:N-t-N-t1} imply that $|\set{N}_{t_1, t}(x)| > A \frac{\log \lfloor \rho t/4 \rfloor}{1 + \log \log \lfloor \rho t/4 \rfloor}$ for sufficiently large $t$.

Using the following lemma, we bound the number that the algorithm enters the monotonization phase between $t_1$~and~$t$:
\begin{lemma} \label{lem:monotone-bound}
For any action $a \in \set{A}$ and three different $u, u', u''$ such that $u < u'< u''$, suppose that the event $\set{E}^{\mnt}_t \cap \{(X_t, A_t) = (x, a)\}$ occurs for all $t \in \{u, u', u''\}$. Then, when $N_u(x)> e$,
\begin{align*}
N_{u''}(x) - N_u(x)  \ge  \frac{N_u(x)}{2 \log N_u(x)} .
\end{align*} 
\end{lemma}
{\bf Proof of Lemma~\ref{lem:monotone-bound}.}
Observe that
selecting action $b$ in the monotonization phase at time $t$ means that 
\begin{align} \label{eq:monotone-observe}
N_t(x,a) \in [\log^2 N_t(x), \log^2 N_t(x) + 1)
\end{align}

From the fact that $u < u' < u''$, we have $N_{u''}(x,a) \ge N_{u}(x,a) + 2$ and thus using \eqref{eq:monotone-observe}:
\begin{align*}
\log^2 N_u(x) + 2  \le N_{u}(x,a) + 2
\le N_{u''}(x,a) 
< \log^2 N_{u''}(x) + 1.
\end{align*} 
We deduce that $\log^2 N_{u''}(x) - \log^2 N_u(x)  > 1$, and conclude that for $N_u(x) > e$,
\begin{align*}
N_{u''}(x) - N_u(x)  \ge \frac{N_u(x)}{2 \log N_u(x)} 
\end{align*}
since the function $\log^2 t$ is concave with derivative $\frac{2\log t}{t}$, i.e., in order to increase $\log^2 t$ by $1$, $t$ should be increased by more than $\left(\frac{2\log t}{t}\right)^{-1}$.
\ep

From Lemma~\ref{lem:monotone-bound}, 
it follows that for sufficiently large $t$, 
\begin{align}
\sum_{b \in \set{A}}\sum_{u = t_1}^{t} \mathbbm{1} [N_t(x) \ge \rho t,  \set{E}^{\mnt}_t, (X_t,A_t) = (x,b)] 
&\le
A \max \left\{3, 
3 (N_t(x)-N_{t_1}(x)) \left(\frac{2 \log N_{t_1}(x)}{N_{t_1}(x)} \right)
\right\} \nonumber \\
&\le
A \max \left\{3, 
3 (t-\lfloor \rho t / 4 \rfloor) \left(\frac{2 \log \lfloor \rho t / 4 \rfloor}{\lfloor \rho t / 4 \rfloor} \right)
\right\} \nonumber \\
&\le 24A \log t  . \label{eq:mnt-bound}
\end{align}
For the first inequality, we apply Lemma~\ref{lem:monotone-bound} with the fact that
as $u$ increases, $N_u (x)$ increases and $\frac{2 \log N_{u}(x)}{N_{u}(x)}$ decreases.
The second inequality is from the definition of $t_1$ and \eqref{eq:N-t-N-t1}.
The last inequality holds for sufficiently large $t$.
We have completed the proof of Lemma~\ref{lem:mono-est-Z1}.
\ep

We return to the proof of Lemma \ref{lem:bound-1}. Lemma ~\ref{lem:mono-est-Z1} establishes \eqref{eq:erg-app-Z1}. Next we provide an upper bound of \eqref{eq:erg-app-Z1}. To this aim, we use the following concentration inequality \citet{Combes2014}: 
 \begin{lemma}  \label{lem:deviation}
Consider any $\phi$, $\pi$, $\epsilon > 0$ with Bernoulli reward distribution. Define ${\cal H}_t$ the $\sigma$-algebra generated by
$( Z_s )_{1 \leq s \leq t}$. Let $\set{B} \subset \mathbb{N}$ be a (random) set of rounds. Assume that there exists a sequence of (random) sets $(\set{B}(s))_{s\ge 1}$ such that (i) $\set{B} \subset \cup_{s \geq 1} \set{B}(s)$, (ii) for all $s\ge 1$ and all $t\in \set{B}(s)$, $N_t(x, a) \ge \epsilon s$, (iii) $|\set{B}(s)| \leq 1$, and (iv) the event $t \in \set{B}(s)$ is ${\cal H}_t$-measurable. Then for all $\zeta > 0$, and $x_1, x, y \in \set{S}$, $a \in \set{A}$,
\begin{align*}
 \sum_{t \geq 1} \Pr^\pi_{\phi \mid x_1}[ t \in \set{B} , | r_t(x, a) - r_\phi ( x,a ) | > \zeta] & \leq 
  \frac{1}{\epsilon \zeta^2} \\
 \sum_{t \geq 1} \Pr^\pi_{\phi \mid x_1}[ t \in \set{B} , | p_t(y \mid x, a) - p_\phi (y \mid x,a ) | > \zeta] &\leq 
  \frac{1}{\epsilon \zeta^2} 
\end{align*}
\end{lemma}
{\bf Proof of Lemma~\ref{lem:deviation}.}
\citet{Combes2014} provides a proof of the first part. Now the occurrence of a transition under action $a$ from state $x$ to state $y$ is a Bernoulli random variable, and hence the second part of the lemma directly follows from the first.
\ep

Let $\set{B}''_t(x,a) := \{\min_{(y,b)\in \set{S}\times \set{A}}N_t(y,b) \ge \frac{\log  t}{(1+\log \log t)^2}, \set{B}'_t(x,a)\}$.
If at time $t \le T$, we have the $s$-th occurrence of $\set{B}''_t(x,a)$, then it follows that $s \le \gamma'_t $ (since $a$ is selected in state $x$ at time $t$, and $N_t(x,a)\le \gamma_t'$), and thus 
\begin{align*}
\min_{(y,b) \in \set{S} \times \set{A}}N_t(y,b)  \ge 
\frac{\log t}{(1+\log \log t)^2}
&\ge
  \frac{1}{ 16S^3 A(1+\gamma) (1+\log\log t)^4 }
 \gamma'_t
  \\
 &\ge
  \frac{1}{16S^3 A(1+\gamma) (1+\log\log T)^4 } s, 
\end{align*}
where the last inequality follows from $t\le T$ and $s\le \gamma_t'$. Thus since $\neg\set{B}_t(\zeta_t)$ holds when $\set{B}''_t(x,a)$ occurs, we deduce that the set of rounds where $\set{B}''_t(x,a)$ occurs satisfies 
\begin{align*}
\{t: \set{B}''_t(x,a) \text{ occurs} \} \subset \cup_{s\ge 1} \cup_{(y,b)\in\set{S}\times\set{A}} \{t: \text{$s$-th occurence of } &\set{B}''_t(x,a), N_t(y,b)\ge \epsilon s, \\
& \| \phi_t(y,b)-\phi(y,b)\| > \zeta_T\},
\end{align*}
where $\epsilon := \frac{1}{16S^3 A (1+\log\log T)^4 }$. Now we apply Lemma~\ref{lem:deviation} to each pair $(y,b)$ with $\zeta=\zeta_T$, and conclude that:

\begin{align*}
 \sum_{t=1}^T   
\Pr^\pi_{\phi | x_1} \left[  
\set{B}''_t(x,a)  \right] 
\le  (SA) \frac{ 16S^3 A (1+\log\log T)^4 }{(\zeta_T)^2} =
16S^4 A^2 (1+\log\log T)^6 =  o (\log T) 
\end{align*}
where 
the factor $SA$ in the inequality is from the union bound over all $(y,b) \in \set{S} \times \set{A}$.
This proves \eqref{eq:bound-1-wts-2} and completes the proof of Lemma~\ref{lem:bound-1}. 
\ep

\subsection{Proof of Lemma~\ref{lem:bound-2}}
\label{sec:pf-bound-2}

Let $\varepsilon_2 := \min_{(x,a) \in \set{S} \times \set{A} : a \notin \set{O} (x;\phi)} (\mathbf{B}^*_\phi h^*_\phi) (x) - 
(\mathbf{B}^a_\phi h^*_\phi) (x) > 0$.
Fix $(x,a) \in \set{S} \times  \set{A}$ such that 
$a \notin \set{O}(x;\phi)$, and $\varepsilon \in (0, \varepsilon_2/5)$
so that 
\begin{align}
(\mathbf{B}^a_\phi h^*_\phi) (x) - (\mathbf{B}^*_\phi h^*_\phi) (x)  \le -5\varepsilon.
\label{eq:choice-of-eps-2}
\end{align}

When $\set{Z}^{(2)}_{t}(x,a; \varepsilon)$ occurs,
we have
\begin{align}
(\mathbf{B}^a_{\phi_t} h^*_\phi) (x) -
(\mathbf{B}^*_\phi h^*_\phi) (x) 
&= 
(\mathbf{B}^a_{\phi_t} h^*_\phi) (x) - (\mathbf{B}^a_{\phi_t} h'_t)(x)
+
(\mathbf{B}^a_{\phi_t} h'_t)(x) -  (\mathbf{B}^*_{\phi} h^*_\phi)(x)  
\nonumber \\
&>
(\mathbf{B}^a_{\phi_t} h^*_\phi) (x) - (\mathbf{B}^a_{\phi_t} h'_t)(x)
- 2 \varepsilon
\nonumber\\
& = 
\left(\sum_{y \in \set{S}} p_t(y \mid x,a) ( h^*_\phi(y)  - h'_t(y) ) \right)
- 2 \varepsilon \nonumber \\
&\ge -3\varepsilon \label{eq:observe-on-Z-2}
\end{align}
where the first inequality stems from the fact that $(\mathbf{B}^a_{\phi_t} h'_t)(x) >  (\mathbf{B}^*_{\phi} h^*_\phi)(x) - 2\varepsilon$ when  $\set{Z}^{(2)}_{t}(x,a; \varepsilon)$ occurs, and the last inequality follows from the fact that $\set{E}_t(\varepsilon)$ holds when  $\set{Z}^{(2)}_{t}(x,a; \varepsilon)$ occurs.

Let $\zeta = \frac{\varepsilon}{S^2}$.
Then, recalling the definition of the event $\set{B}_t (x,a ; \zeta) := \{\|  \phi_t(x,a) - \phi(x,a) \| \le \zeta \}$, when $\set{B}_t (x,a ; \zeta)$ occurs, 
we have
\begin{align}
|(\mathbf{B}^a_{\phi} h^*_\phi) (x)  - (\mathbf{B}^a_{\phi_t} h^*_\phi) (x) |
&\le  |r_t(x,a) - r_\phi (x,a)| + H_{\phi} \sum_{y \in \set{S}} |p_t(y \mid x,a) - p_\phi (y  \mid x,a) | \nonumber \\
&\le  |r_t(x,a) - r_\phi (x,a)| + S^2 \max_{y \in \set{S}} |p_t(y \mid x,a) - p_\phi (y  \mid x,a) | \nonumber \\
&\le S^2 \|\phi_t(x,a) - \phi(x,a) \| \nonumber \\
&\le \varepsilon \label{eq:observe-on-B}
\end{align}
where for the second inequality, we used $0 \le H_\phi \le S$.

Now, we can deduce that
the events $\set{Z}^{(2)}_t(x,a ; \varepsilon)$ and $\set{B}_t (x,a ; \zeta)$
cannot occur at the same time, i.e.,
\begin{align} \label{eq:B-Z-exclude}
\Pr^\pi_{\phi \mid x_1}\left[ \set{Z}^{(2)}_t(x,a; \varepsilon), \set{B}_t (x,a ;\zeta) \right] = 0.
\end{align}
 Indeed, when $\set{Z}^{(2)}_t(x,a ; \varepsilon) \cap \set{B}_t (x,a ; \zeta)$ occurs, \eqref{eq:observe-on-Z-2} and 
\eqref{eq:observe-on-B} imply 
\begin{align*}
(\mathbf{B}^a_{\phi} h^*_\phi) (x) -
(\mathbf{B}^*_\phi h^*_\phi) (x) 
&=(\mathbf{B}^a_{\phi_t} h^*_\phi) (x) -
(\mathbf{B}^*_\phi h^*_\phi) (x) 
- \left((\mathbf{B}^a_{\phi} h^*_\phi) (x)  - (\mathbf{B}^a_{\phi_t} h^*_\phi) (x) \right) \\
&\ge
(\mathbf{B}^a_{\phi_t} h^*_\phi) (x) -
(\mathbf{B}^*_\phi h^*_\phi) (x) 
- |(\mathbf{B}^a_{\phi} h^*_\phi) (x)  - (\mathbf{B}^a_{\phi_t} h^*_\phi) (x) | \\
& \ge -4 \varepsilon > -5 \varepsilon
\end{align*}
which contradicts \eqref{eq:choice-of-eps-2} for our choice of $\varepsilon$ , i.e., $\varepsilon \in (0, \varepsilon_2/5)$.

Hence, to complete the proof, it is sufficient to show that
\begin{align} \label{eq:bound-2-wts}
\sum_{t=1}^T \Pr^\pi_{\phi \mid x_1} [(X_t, A_t) = (x,a), \neg\set{B}_t (x,a ;\zeta)]
= O(1) 
\end{align} 
as we have the following bound:
\begin{align*}
\sum_{t =1}^T \Pr^\pi_{\phi \mid x_1}\left[ \set{Z}^{(2)}_t(x,a; \varepsilon) \right]
&= \sum_{t =1}^T \Pr^\pi_{\phi \mid x_1}\left[ \set{Z}^{(2)}_t(x,a; \varepsilon), \neg\set{B}_t (x,a ;\zeta) \right]\\
&\le \sum_{t=1}^T \Pr^\pi_{\phi \mid x_1} [(X_t, A_t) = (x,a), \neg\set{B}_t (x,a ;\zeta)]
\end{align*}
where the equality follows from \eqref{eq:B-Z-exclude}. \eqref{eq:bound-2-wts} is obtained by applying Lemma~\ref{lem:deviation}
with $\{(X_t, A_t)  = (x,a)\}$, $1$ and
$\frac{\varepsilon}{S^2}$ for $\set{B}$, $\epsilon$ and $\zeta$, respectively.
This complete the proof of Lemma~\ref{lem:bound-2}.

\ep


\subsection{Proof of Lemma~\ref{lem:bound-3}}
\label{sec:pf-bound-3}


Recall that:
\begin{align*}
\set{E}_t(\varepsilon) :=
\left\{ 
\Pi^*(\phi'_t) \subseteq \Pi^*(\phi)
~\text{and}~
|r_t(x,a) - r_{\phi}(x,a) | + |h'_t (x) - h^*_\phi(x)| \le \varepsilon 
~\forall x \in \set{S}, \forall a \in \set{O}(x;\phi'_t)
\right\} .
\end{align*}
Hence when $\set{E}_t(\varepsilon)$ occurs, ($i$) the estimation of the bias function in the restricted MDP $\phi(\set{C}_t)$ is accurate and ($ii$) the restricted MDP includes the optimal policies of $\phi$. We first focus on the accuracy of the estimated bias function, and then show that the gain of the restricted MDP $\phi(\set{C}_t)$ is monotone increasing and that it eventually includes an optimal policy for the (unrestricted) MDP. 

{\bf Estimation error in bias function. } 
We begin with some useful notations.
Let $K:= A^S$ be the number of all the possible fixed policies.
Fix $\beta \in \left(0, \frac{1}{K + 1}\right)$. 
For sufficiently large $t > \frac{1}{\frac{1}{K +1} - \beta}$,
divide the time interval from $1$ to $t$
into $(K+1)$ subintervals $\set{I}^t_0, \set{I}^t_1, \dots, \set{I}^t_{K}$
such that $\set{I}^t_k := \{u \in \mathbb{N} : i^t_k \le u < i^t_{k+1} \}$
where $i^t_0 :=1$ and $i^t_k := t +1- (K +1- k) \lfloor \frac{t}{K+1}\rfloor$
for $k \in \{1, ...,  K+1\}$.
Then, it is easy to check that for each $k \in \{0, ..., K\}$, 
\begin{align*}
|\set{I}^t_k| = i^t_{k+1} - i^t_k > \beta t .
\end{align*}
Indeed, for $k = 0$, $i^t_1 - i^t_0 = t - K \lfloor \frac{t}{K+1} \rfloor \ge \frac{t}{K+1} > \beta t$,
and for $k \in [1, K]$, $i^t_{k+1} - i^t_k =  \lfloor \frac{t}{K+1} \rfloor
 \ge \frac{t}{K+1} > \beta t$ as $t > \frac{1}{\frac{1}{K +1} - \beta}$, i.e., 
each subinterval length grows linearly with respect to $t$.

For $k \in [0, K]$, $x \in \set{S}$ and $a \in \set{A}$,
let $N^t_k (x) := N_{i^t_{k+1}} (x) -  N_{i^t_{k}} (x) $
and $N^t_k (x, a) := N_{i^t_{k+1}} (x, a) -  N_{i^t_{k}} (x, a)$.
Using $\rho > 0$ in \eqref{eq:erg}, for $\zeta > 0$, define an event $\set{D}_t (\zeta)$ as
\begin{align} \label{eq:define-D_t}
\set{D}_t (\zeta) := \set{D}'_t \cap \set{E}'_t(\zeta)
\end{align}
where we let
\begin{align*}
\set{D}'_t &:= 
\left\{
N^t_{k}(x) > \rho \beta t, 
\forall x \in \set{S}, \forall k \in [0, K]
\right\}  \\
\set{E}'_t (\zeta) &:= 
\left\{
\| \phi'_u - \phi(\set{C}_u) \| \le  \zeta
~
\forall u \in [i^t_1, t]
\right\} .
\end{align*}
When $\set{D}_t(\zeta)$ occurs, then in each subinterval, each state is linearly visited,
and after the first subinterval, the estimation on the restricted MDP is accurate, i.e., $\phi(\set{C}_t) \simeq \phi_t(\set{C}_t)$.
Note that $\set{E}_t(\varepsilon)$ bounds the error in the estimated gain and bias functions.
Hence, we establishy the correspondence between $\zeta$ in $\set{D}_t(\zeta)$ and $\varepsilon$ in $\set{E}_t(\varepsilon)$
using the continuity of the gain and bias functions in $\phi$:  
\begin{lemma}  \label{lem:translate-eps-to-zeta}
Consider an ergodic MDP $\phi$ with Bernoulli rewards.
Then, for $\varepsilon >0$.
there exists $\zeta_0 = \zeta_0(\varepsilon, \phi) >0 $ such that
for any $\zeta \in (0, \zeta_0)$, policy $f \in \Pi_D$ and MDP $\psi$,
if $\|\psi - \phi \| \le \zeta$ and $\psi \ll \phi$, then $\psi$ is ergodic, $|g^f_\psi - g^f_\phi | \le \varepsilon$  and $\|h^f_\psi - h^f_\phi \| \le \varepsilon$.
\end{lemma}
The proof of Lemma~\ref{lem:translate-eps-to-zeta} is in Section~\ref{sec:pf-translate-eps-to-zeta}.  
Observe that on the event $\set{D}_t (\zeta)$, for $u \ge i^t_1$, every state is visited more than
$\rho\beta t$, i.e., $\log^2 N_u (x) \ge \log^2 \rho \beta t \ge 1$ for all $x\in \set{S}$ and sufficiently large $t$, and thus, for all $(x,a) \in \set{S} \times \set{A}$ such that $a \in \set{C}_u(x)$,
 $\phi_u(x,a)$ is indeed the estimation of $\phi(x,a)$, i.e., $\phi'_u = \phi_u(\set{C}_u) \ll \phi(\set{C}_u)$.
Then, using Lemma~\ref{lem:translate-eps-to-zeta},
it follows that there exists constant $t_0>0$ such that for $t > t_0$
and $\zeta \in(0, \min\{\zeta_0(\varepsilon/2, \phi), \varepsilon / 2\})$, 
\begin{align}
\set{D}_t(\zeta) \subseteq 
 \{
\text{$\phi'_u$ is ergodic}~\forall u \in [i^t_1, t]
\}
\cap
\set{E}''_t (\varepsilon)  \label{eq:on-D-connect}
\end{align}
where 
\begin{align*}
\set{E}''_t (\varepsilon) := 
\left\{
| r_{\phi'_u}(x, f(x)) - r_{\phi} (x, f(x)) |
+
| h^f_{\phi'_u}(x)- h^f_{\phi}(x) | \le  \varepsilon
~\forall u \in [i^t_1, t], \forall f \in  \Pi_{D}(\set{C}_u),
\forall x \in \set{S}
\right\},
\end{align*}
and where for restriction $\set{C} : \set{S} \twoheadrightarrow \set{A}$,
we denote by $\Pi_{D} (\set{C})$ the set of all the possible deterministic policies on the restricted MDP $\phi(\set{C})$.

{\bf Monotone improvement. } 
Based on \eqref{eq:on-D-connect}, we can identify instrumental properties of DEL algorithm when $\set{D}_t(\zeta)$ occurs:
\begin{lemma} \label{lem:D-implies}
For structure $\Phi$ with Bernoulli rewards and an ergodic MDP $\phi \in \Phi$,
consider $\pi = \textnormal{DEL}$.
There exists $\zeta_1 > 0$ and $t_1 > 0$ such that
for any $\zeta \in (0, \zeta_1)$ and $t > t_1$,
the occurrence of the event $\set{D}_t(\zeta)$ implies that 
\begin{align}
\Pi^*(\phi'_u)  \subseteq \Pi^*(\phi(\set{C}_u)),
 \quad \text{and}\quad
g^*_{u+1}  \ge g^*_{u}, \quad
\forall u \in [i^t_1, t]
\label{eq:D-implies}
\end{align}
where we denote by $g^*_{u} := g^*_{\phi(\set{C}_u)}$ 
and $h^*_{u} := h^*_{\phi(\set{C}_u)}$  the optimal gain and bias functions, respectively,
on the restricted MDP $\phi(\set{C}_u)$ with true parameter $\phi$.
\end{lemma}
The proof of Lemma~\ref{lem:D-implies} is presented in Section~\ref{sec:pf-D-implies}. 

Define the event
\begin{align*}
\set{M}_t := \left\{g^*_{i^t_{k+1}} > g^*_{i^t_{k}}~\forall k \in [1, K]
 ~\text{or}~
g^*_{i^t_{k+1}} = g^*_{i^t_{k}} = g^*_\phi~\text{for some $k\in [1, K]$}
\right\}.
\end{align*}
Then, by selecting $\zeta$ as in in Lemma~\ref{lem:D-implies} and \eqref{eq:on-D-connect},
we can connect the events $\set{M}_t $ and $ \set{D}_t (\zeta)$ to the event $\set{E}_t(\varepsilon)$
as follows: for $\zeta \in (0, \min\{\zeta_0(\varepsilon/2, \phi), \varepsilon/2, \zeta_1\})$ and sufficiently large $t > t_1$,
\begin{align}  \label{eq:MD-implies-E}
\set{M}_t \cap \set{D}_t (\zeta) \subseteq \set{E}_t(\varepsilon).
\end{align}
On the event $\set{M}_t$, there must exists $k\in[1, K+1]$ such that $g^*_{i^t_k} = g^*_\phi$
since the number $K$ of subintervals is the number of all the possible policy $\Pi^* $.
In addition, for such a $k\in[1, K+1]$, on the event $\set{D}_t (\zeta)$,
it follows from Lemma~\ref{lem:D-implies} that for all $u \in [ i^t_k, t]$, $g^*_{i^t_k} = g^*_\phi \le g^*_u \le g^*_t$, i.e., $g^*_t =g^*_\phi$ and thus $\Pi^*(\phi(\set{C}_t)) \subseteq \Pi^*(\phi)$.
Therefore, when both of the events $\set{M}_t$ and $\set{D}_t (\zeta)$ occur,
\begin{align*}
\Pi^*(\phi) \supseteq \Pi^*(\phi(\set{C}_t))  \supseteq \Pi^*(\phi'_t) 
\end{align*}
(again thanks to Lemma~\ref{lem:D-implies}).
Then, we indeed get $\set{M}_t \cap \set{D}_t (\zeta) \subseteq \set{E}_t(\varepsilon)$: the ergodicity of $\phi'_t$ guaranteed from  \eqref{eq:on-D-connect} implies 
that the optimal bias function $h'_t$ of $\phi'_t$ is  unique, and the event $\set{E}''_t(\varepsilon)$ in \eqref{eq:on-D-connect}
always occurs on the event $\set{D}_t (\zeta)$. Thus the estimated bias function $h'_t$ is close to $h^*_{\phi}$.

Using \eqref{eq:MD-implies-E} and \eqref{eq:define-D_t}, for small enough
$\zeta \in (0, \min\{\zeta_0(\varepsilon/2, \phi), \varepsilon/2, \zeta_1\})$
and for large enough $t > 0$, we get
\begin{align}
 \Pr^\pi_{\phi \mid x_1} \left[\neg \set{E}_t(\varepsilon) \right]  
&\le 
+ \Pr^\pi_{\phi \mid x_1} \left[\neg\set{D}_t(\zeta) \right] 
+ \Pr^\pi_{\phi \mid x_1} \left[\set{D}_t(\zeta), \neg\set{M}_t \right] 
 \nonumber \\
&\le O(1)
+ \Pr^\pi_{\phi \mid x_1} \left[\neg\set{D}'_t \right]  
+ \Pr^\pi_{\phi \mid x_1} \left[\set{D}'_t, \neg\set{E}'_t(\zeta)\right]  
 +  \Pr^\pi_{\phi \mid x_1} \left[\set{D}_t(\zeta), \neg\set{M}_t \right] 
\label{eq:bound-3-final}
\end{align}
where the first and last inequalities are from \eqref{eq:MD-implies-E}
and \eqref{eq:define-D_t}, respectively.
To complete the proof of Lemma~\ref{lem:bound-3}, we provide upper bounds of each term in the r.h.s. of \eqref{eq:bound-3-final}. the first term can be easily bounded. Indeed, using \eqref{eq:erg} and a union bound, we get for $t$ sufficiently,
\begin{align}
\Pr^\pi_{\phi \mid x_1} [\neg\set{D}'_t] 
\le 
\sum_{x \in \set{S}}
\sum_{k \in [0, K]}
\Pr^\pi_{\phi \mid x_1} [N^t_{k}(x) \le \rho \beta t]  = o(1/t)
\label{eq:bound-3-first-sum}
\end{align}
where the last equality is from \eqref{eq:erg} conditioned on $X_{i^t_k}$ for each $k$.
 
Lemma \ref{lem:difficult-bound-3} below deals with the last term. 

\begin{lemma} \label{lem:difficult-bound-3}
For structure $\Phi$ with Bernoulli rewards and an ergodic MDP $\phi \in \Phi$,
consider $\pi = \textnormal{DEL}$.
Suppose $\phi$ is in the interior of $\Phi$, i.e., 
there exists a constant $\zeta_0 >0$ such that for any $\zeta \in (0, \zeta_0)$,
$\psi \in \Phi$ if $\|\phi - \psi\| \le \zeta$.
There exists $\zeta_2 > 0 $ such that for $\zeta \in (0, \zeta_2)$,
\begin{align*}
\Pr^\pi_{\phi \mid x_1} \left[\set{D}_T(\zeta), \neg\set{M}_T \right]  = o(1/T)
\quad \text{as $T \to \infty$}.
\end{align*}
\end{lemma}
We provide the proof of Lemma~\ref{lem:difficult-bound-3} in Section~\ref{sec:pf-difficult-bound-3}. There, the assumption that $\phi$ is in the interior of $\Phi$ plays an important role when studying the behavior of the algorithm in the exploitation phase.

To bound the second term in the r.h.s. of \eqref{eq:bound-3-final}, we use the following concentration inequality:
\begin{lemma} \label{lem:concen-exp}
Consider any $\pi$ and $x_1 \in \set{S}$.
There exist $C_0, c_0, u_0 > 0$ such that for any $(x,a) \in \set{S} \times \set{A}$
and $u \ge u_0$,
\begin{align*}
\Pr^\pi_{\phi \mid x_1} [ |\phi_t (x,a) -  \phi (x,a)| >  \zeta, N_t (x,a) = u] \le  C_0 e^{- c_0 u } .
\end{align*}
\end{lemma}
{\bf Proof of Lemma~\ref{lem:concen-exp}. }
The proof is immediate from Lemma~4(i)
in \cite{burnetas1997optimal}, which is an application of Cramer's theorem for estimating Bernoulli random variables. 
Let $\hat{\phi}_t(x,a)$ be the estimator of $\phi(x,a)$ from $t$ i.i.d. reward and transition samples when action $a$ is selected in state $x$. From Lemma~4(i) in \cite{burnetas1997optimal}, there are positive constants $C(x,a)$, $c(x,a)$, and $u_0(x,a)$ (which may depend on $(x,a)$), such that for $u \ge u_0(x,a)$,
\begin{align*}
\Pr^\pi_{\phi \mid x_1} [ |\phi_t (x,a) -  \phi (x,a)| >  \zeta, N_t (x,a) = u] 
& \le \Pr [ |\hat{\phi}_u (x,a) -  \phi (x,a)| >  \zeta]  \\
& \le  C_0(x,a)  e^{- c_0(x,a) u }.
\end{align*}
We complete the proof by taking $C_0 := \max_{(x,a) \in \set{S} \times \set{A}} C_0(x,a)$,
$c_0 := \min_{(x,a) \in \set{S} \times \set{A}} c_0(x,a)$, and
$u_0 := \max_{(x,a) \in \set{S} \times \set{A}} u_0(x,a)$.
\ep

Now observe that:
\begin{align*} 
&\Pr^\pi_{\phi \mid x_1} [\set{D}'_t, \neg\set{E}'_t(\zeta)] \\
&= \Pr^\pi_{\phi \mid x_1} [\set{D}'_t, 
\| \phi_u(x,a) - \phi(x,a) \| > \zeta, 
\text{for some}~ u \in [ i^t_1,t],  x \in \set{S}, a \in \set{C}_u (x)
]  \\
&\le 
\sum_{u=i^t_1}^t
\sum_{x \in \set{S}}
\sum_{a \in \set{C}_u(x)}
\Pr^\pi_{\phi \mid x_1} [\set{D}'_t, 
\| \phi_u(x,a) - \phi(x,a) \| > \zeta]  \\
&\le
\sum_{u=i^t_1}^t
\sum_{x \in \set{S}}
\sum_{a \in \set{C}_u(x)}
\Pr^\pi_{\phi \mid x_1} [ 
\| \phi_u(x,a) - \phi(x,a) \| > \zeta, N_u(x,a) \ge \log^2 N_u(x), \rho \beta t \le N_u(x) \le u]  \\
&\le
\sum_{u=i^t_1}^t
\sum_{x \in \set{S}}
\sum_{a \in \set{C}_u(x)}
\sum_{u' = \rho \beta t}^u
\sum_{u'' = \log^2 {u'}}^{u'}
\Pr^\pi_{\phi \mid x_1} [
\| \phi_u(x,a) - \phi(x,a) \| > \zeta, N_u(x,a) =  u''] 
\end{align*}
where the second inequality follows from the definition of $\set{C}_u(x)$ and the fact that
on the event $\set{D}'_t$,  $ \rho \beta t \le N_0^t(x) \le N_u(x)$ for $u \in[ i^t_1, t]$.
Then, applying Lemma~\ref{lem:concen-exp}, 
we have
\begin{align} 
\Pr^\pi_{\phi \mid x_1} [\set{D}'_t, \neg\set{E}'_t(\zeta)] 
&\le
\sum_{u=i^t_1}^t
\sum_{x \in \set{S}}
\sum_{a \in \set{C}_u(x)}
\sum_{u' = \rho \beta t}^u
\sum_{u'' = \log^2 u'}^{\infty}
C_0 e^{-c_0 u''} \nonumber \\
&\le SA C_0  t^2 \frac{e^{-c_0 \log^2 (\rho \beta t)}}{1-e^{-c_0}} \nonumber  \\
& = \frac{SA C_0}{{1-e^{-c_0}}}  t^2 e^{-c_0 (\log^2 (\rho \beta) + \log t (\log t+2 \log (\rho \beta)))}
\nonumber  \\
& = 
\frac{SA C_0}{{1-e^{-c_0}}} 
e^{-c_0 \log^2 (\rho \beta) }
 t^{2 -  2c_0  \log (\rho \beta) - c_0 \log t} = o(1/t) . \label{eq:bound-3-second-sum}
\end{align}
Combining Lemma~\ref{lem:difficult-bound-3}, \eqref{eq:bound-3-first-sum}
and \eqref{eq:bound-3-second-sum}
to \eqref{eq:bound-3-final}, we complete the proof of Lemma~\ref{lem:bound-3}.

\ep

\subsubsection{Proof of Lemma~\ref{lem:translate-eps-to-zeta}}
\label{sec:pf-translate-eps-to-zeta}

Define two strictly positive constants:
\begin{align*}
\zeta_r (\phi) &:= 
\min \left\{
\frac{r_{\phi}(x,a)}{2} : \forall x \in \set{S}, \forall a \in \set{A} \text{ s.t. } r_{\phi}( x,a) > 0 
\right\} 
\end{align*}

\begin{align*}
 \zeta_p (\phi) &:= 
\min \left\{
\frac{p_{\phi}(y \mid x,a)}{2} : \forall x,y \in \set{S}, \forall a \in \set{A} \text{ s.t. } p_{\phi}(y\mid x,a) > 0  \right\}. 
\end{align*} 
Then, it is straightforward to show that $\phi \ll \psi$ if $\|\psi - \phi\| < \min \{\zeta_r (\phi),  \zeta_p (\phi)\}$ since for any $x,y \in \set{S}$ and $a \in \set{A}$,
$p_{\phi}(y \mid x,a) > 0$ implies that $p_{\psi}(y \mid x,a) \ge p_{\phi}(y \mid x,a)/2 > 0$,
and
$r_{\phi}(x,a) > 0$ implies that $r_{\psi}(x,a) \ge r_{\phi}(x,a)/2 > 0$.
Therefore, for sufficiently small $\zeta_0 \le \min \{\zeta_r (\phi),  \zeta_p (\phi)\}$, 
the above observation  and the assumption that $\psi \ll \phi$ ensure the mutual absolute continuity between $\phi$ and $\psi$
and thus the ergodicity of $\psi$.

Now, we focus on the continuity of gain and bias functions for given policy $f$.
For notational convenience, let $g^f_\phi$ (resp. $g^f_\psi$) and 
$h^f_\phi$ (resp. $h^f_\psi$)
denote the (column) vector of gain and bias functions, respectively, under $\phi$ (resp. $\psi$).
Let $P^f_{\phi}$ (resp. $P^f_\psi$) and $r^f_\phi$ (resp. $r^f_\phi)$ are the transition matrix and reward vector w.r.t. policy $f$ under $\phi$ (resp. $\psi$), respectively.
Then, we can write the policy evaluation equations of stationary policy $f$ under $\phi$ and $\psi$ as vector and matrix multiplications, c.f., \cite{puterman1994markov}:
\begin{align*}
g^f_\phi &= P^f_\phi g^f_\phi  \\
h^f_\phi &= r^f_\phi - g^f_\phi + P^f_\phi h^f_\phi. 
\end{align*}
Similarly $g^f_\psi = P^f_\psi g^f_\psi$ and $h^f_\psi = r^f_\psi - g^f_\psi + P^f_\psi h^f_\psi$.
Since both $\phi$ and $\psi$ are ergodic, by forcing $h^f_{\phi} (x_1) = h^f_{\psi} (x_1) = 0$ for some $x_1 \in \set{S}$, the bias functions $h^f_{\phi}$ and $h^f_{\psi}$ can be uniquely defined. 
Let $D^f := P^f_\phi - P^f_\psi$ and $d^f := h^f_\phi - h^f_\psi$.
Then, $\| D^f \| \le S\zeta$ where $\|\cdot \|$ is the max norm.
Noting that the ergodicity of $\phi$ and $\psi$ further provides the invertibility of $I - P^f_\phi$ and $I - P^f_\psi$.
A basic linear algebra, c.f., Lemma~7 in \cite{burnetas1997optimal}, leads to that for any $\varepsilon  > 0$, $ \|d^f\| \le \varepsilon$ if
\begin{align*}
\| D^f\| \le \frac{\varepsilon}{\|(I-P^f_\phi)^{-1} \| (\|h^f_\phi \|+ \varepsilon)}
\end{align*}
where the upper bound is independent of $\psi$.
From the above continuity of $h^f_{\psi}$ (and thus that of $g^f_\psi$) 
with respect to $\psi$ at $\phi$, we can find $\zeta_0(f, \varepsilon, \phi) > 0$
such that for any $\psi$, $|g^f_\psi - g^f_\phi| \le \varepsilon$ and
$\|h^f_\psi - h^f_\phi \| \le \varepsilon$ if 
$\|\psi -\phi\| \le \zeta_0(f, \varepsilon,\phi)  \le \min \{\zeta_r(\phi), \zeta_p(\phi)\}$.
Noting the arbitrary choice of $f \in \Pi_D$,
we conclude the proof of Lemma~\ref{lem:translate-eps-to-zeta}
by taking $\zeta_0 (\varepsilon, \phi) = \min_{f \in \Pi_D} \zeta_0(f, \varepsilon,\phi)$.
\ep

\subsubsection{Proof of Lemma~\ref{lem:D-implies}}
\label{sec:pf-D-implies}

Let $\varepsilon_1 := \min \{ |g^f_\phi - g^{f'}_\phi| :  f, f' \in \Pi_D,g^f_\phi  \neq g^{f'}_\phi \} > 0$. Let $\zeta_1 := \min \{\zeta_0( \frac{\varepsilon_1}{2S}, \phi), \frac{\varepsilon_1}{2S}\}$ and consider $t$ sufficiently large, i.e., $t  >t_0$.
For $\zeta \in  (0,  \zeta_1)$, assume that the event $\set{D}_t(\zeta)$ occurs.

{\bf Proof of the first part of \eqref{eq:D-implies}.}
Then, 
for any $u \in [i^t_1, t]$ and $f \in \Pi_D(\set{C}_u)$,
it follows from \eqref{eq:on-D-connect} that for any $x \in \set{S}$, 
\begin{align}
|g^f_{\phi'_u} -g^f_{\phi}|  
& =
|(\mathbf{B}^f_{\phi'_u}h^f_{\phi'_u})(x) -(\mathbf{B}^f_{\phi}h^f_{\phi}) (x)| \nonumber \\
& \le | r_{\phi'_u}(x, f(x)) - r_{\phi} (x, f(x)) |
+  \sum_{y \in \set{S}}  | h^f_{\phi'_u}(y) - h^f_{\phi}(y) |  \nonumber \\
& \le S \frac{\varepsilon_1}{2S} =  \frac{\varepsilon_1}{2}   \label{eq:D-implies-1}
\end{align}
where the last inequality stems from the definition of $\set{E}''_t(\frac{\varepsilon_1}{2S})$
in \eqref{eq:on-D-connect}.
Then, for any $u \in [i^t_1, t]$, $f \in  \Pi^*(\phi'_u)$, and $f' \in \Pi_D(\set{C}_u)$, we have:
\begin{align*}
g^f_{\phi} ~\ge~ g^f_{\phi'_u} - \frac{\varepsilon_1}{2} 
~\ge~ g^{f'}_{\phi'_u} - \frac{\varepsilon_1}{2} 
~\ge~ g^{f'}_{\phi} - \varepsilon_1
\end{align*}
where the first and last inequalities stem from \eqref{eq:D-implies-1},
and the second inequality is deduced from the optimality of $f$ under $\phi'_u$.
Noting that $f, f' \in \Pi_D(\set{C}_u)$,
it follows that
$g^f_{\phi(\set{C}_u)} = g^f_{\phi} \ge  g^{f'}_{\phi} = g^{f'}_{\phi(\set{C}_u)}$. 
Hence $f$ is optimal under $\phi(\set{C}_u)$ (the choice of $f' \in \Pi_D(\set{C}_u)$ is arbitrary). This completes the proof of the first part in \eqref{eq:D-implies}.

{\bf Proof of the second part of \eqref{eq:D-implies}.}
Fix $u \in [i^t_1, t]$.
Assume that
\begin{align} \label{eq:D-implies-assum}
\Pi_D(\set{C}_{u+1}) \cap  \Pi^*(\phi'_u) \neq \emptyset .
\end{align} 
Then, from the first part of \eqref{eq:D-implies}, we deduce that:
\begin{align*}
\Pi_D(\set{C}_{u+1}) \cap  \Pi^*(\phi'_u) 
~\subseteq~ \Pi_D(\set{C}_{u+1}) \cap  \Pi^*(\phi(\set{C}_u)).
\end{align*}
Combining this with the assumption \eqref{eq:D-implies-assum}, we get that $\Pi_D(\set{C}_{u+1}) \cap  \Pi^*(\phi(\set{C}_u))\neq\emptyset$, which implies that $g^*_{u+1}  \ge g^*_{u}$. 
It remains to prove \eqref{eq:D-implies-assum}.

Let $x = X_u$. We first show that: 
\begin{equation}\label{eq:qqq}
\set{C}_{u+1} (x) \cap \set{O} (x ; \phi'_u) \neq \emptyset.
\end{equation}

If the algorithm enters the monotonization phase, i.e., the event $\set{E}^\mnt_u$ occurs, then it selects action $a=A_u\in {\cal C}_u(x) \cap \set{O} (x ; \phi'_u)$. We deduce that: 
\begin{align*}
N_u(x, a) \ge \log^2( N_u(x)), \quad
N_{u+1}(x, a)  = N_{u}(x, a)  + 1,  \quad\text{and}\quad 
N_{u+1}(x)  = N_{u}(x)  + 1
\end{align*}
Thus, using the fact that $\log^2 (n) +1 > \log^2 (n+1)$, we obtain
\begin{align}
N_{u+1}(x,a)  \ge \log^2 (N_u (x)) +1  \ge \log^2 (N_u (x) +1) 
= \log^2 (N_{u+1} (x)). \label{eq:monotone-log-trick}
\end{align}
We have shown that $a\in \set{C}_{u+1}(x)$ and thus $a \in \set{C}_{u+1} (x) \cap \set{O} (x ; \phi'_u) \neq \emptyset$.

In case that the event $\set{E}^\mnt_u$ does not occur, 
there must exist an action $a \in \set{O}(x; \phi'_u)$ such that 
$N_u(x, a) \ge \log^2( N_u(x)) + 1$. Hence, as for \eqref{eq:monotone-log-trick}, we get: 
\begin{align*}
N_{u+1}(x,a) \ge N_{u}(x,a)  \ge \log^2 (N_u (x)) +1  \ge \log^2 (N_u (x) +1) 
= \log^2 (N_{u+1} (x)) 
\end{align*}
which implies $a \in \set{C}_{u+1} (x) \cap \set{O} (x ; \phi'_u) \neq \emptyset$. 

Now \eqref{eq:qqq} implies \eqref{eq:D-implies-assum} (since for any $y \in \set{S}$ such that $y \neq x$, $\set{C}_u (y) = \set{C}_{u+1} (y)$). This completes the proof of the second part in \eqref{eq:D-implies} and that of  Lemma~\ref{lem:D-implies}.
\ep

\subsubsection{Proof of Lemma~\ref{lem:difficult-bound-3}}
\label{sec:pf-difficult-bound-3}

We will show that for small enough $\zeta >0$,
\begin{align*}
\Pr^\pi_{\phi \mid x_1} \left[\set{D}_t(\zeta), \neg\set{M}_t \right] = o (1/t)
\end{align*}
where we recall
\begin{align*}
\set{M}_t := \left\{g^*_{i^t_{k+1}} > g^*_{i^t_{k}}~\forall k \in [1, K]
 ~\text{or}~
g^*_{i^t_{k+1}} = g^*_{i^t_{k}} = g^*_\phi~\text{for some $k\in [1, K]$}
\right\}.
\end{align*}
For $x \in\set{S}$ and restriction $\set{C}: \set{S} \twoheadrightarrow \set{A}$,
define 
\begin{align*}
\set{A}^+ (x; \phi, \set{C}) &:= 
\{a \in \set{A} : (\mathbf{B}^a_{\phi} h^*_{\phi(\set{C})}) (x) > 
(\mathbf{B}^*_{\phi(\set{C})} h^*_{\phi(\set{C})})(x) \}
\end{align*} 
as the set of actions that improve the optimal policy of the restricted MDP $\phi(\set{C})$ at state $x$.
If $g^*_{\phi(\set{C})} < g^*_{\phi}$, then there must exist a state $x$ 
with non-empty $\set{A}^+ (x; \phi, \set{C}) $.
Let $\varepsilon_2 :=
 \min \{  (\mathbf{B}^a_{\phi} h^f_{\phi})(x) -  (\mathbf{B}^f_{\phi} h^f_{\phi})(x) :
 f \in \Pi_D, x \in \set{S}, a \in \set{A}^+ (x; \phi, \{f\})  \neq \emptyset \}$.
 Note that $\varepsilon_2 >0$.

Define an event 
\begin{align*}
\set{M}'_t := 
\{\set{E}^\xpt_u ,  \set{A}^+ (X_u; \phi, \set{C}_u) \neq \emptyset, \exists u \in [i^t_1, t]\} .
\end{align*}
Then, we obtain
\begin{align*}
\Pr^\pi_{\phi \mid x_1} \left[\set{D}_t(\zeta), \neg\set{M}_t \right]
\le
\Pr^\pi_{\phi \mid x_1} \left[\set{D}_t(\zeta), \neg\set{M}_t, \neg\set{M}'_t  \right]
+  \Pr^\pi_{\phi \mid x_1} \left[\set{D}_t(\zeta), \set{M}'_t \right].
\end{align*}
We first focus on the last term in the above.
Let $\zeta \le \min \{\zeta_0 (\frac{\varepsilon_2}{3S}, \phi), \frac{\varepsilon_2}{3S}, \zeta_0\}$ where $\zeta_0$ is taken from the assumption that
$\phi$ is in the interior of $\Phi$, and $t \ge \frac{1}{\beta} e^{e^{\varepsilon_2 / 3}}$
so that $\zeta_u \le \varepsilon_2 /3$ for any $u \ge i^t_1 \ge \beta t$.

Suppose that for $u \in [i^t_1, t]$
and $x \in \set{S}$,
the events 
$\set{D}_t(\zeta)$ and 
$\{X_u= x, \set{E}^\xpt_u, \set{A}^+ (x; \phi, \set{C}_u) \neq \emptyset\}$
occur.
From Lemma~\ref{lem:D-implies},
it directly follows that 
$\set{O}(x;\phi'_u) \subseteq \set{O}(x; \phi, \set{C}_u)$.
By the definition of the improving action set,
$\set{O}(x; \phi, \set{C}_u) \cap \set{A}^+ (x; \phi, \set{C}_u) = \emptyset$
and thus $\set{O}(x;\phi'_u) \cap \set{A}^+ (x; \phi, \set{C}_u) = \emptyset$.
Construct $\psi_u$ such that for each $(y,b) \in \set{S} \times \set{A}$, 
\begin{align*}
\psi_u (y, b) =  
\begin{cases}
\phi_u(y, b)  & \text{if $b \in \set{O}(x; \phi'_u)$},  \\ 
\phi(y, b)  & \text{otherwise}. 
\end{cases}
\end{align*}
Note that $\psi_u (\set{C}_u) = \phi'_u$ and thus
$\|\psi_u -\phi\| \le  \|\phi'_u - \phi (\set{C}_u)\|  \le \zeta_0 $.
This implies  $\psi_u \in \Phi$ since $\phi$ is an interior point of $\Phi$.
For any $a \in \set{A}^+ (x; \phi, \set{C}_u) \neq \emptyset$, 
we get $\delta^*(x,a;\psi_u, \set{C}_u, \zeta_u) = 0$ as:
\begin{align}
\delta^*(x,a;\psi_u, \set{C}_u) 
= (\mathbf{B}^*_{\phi'_u} h'_{u}) (x) -
(\mathbf{B}^a_{\psi_u} h'_{u}) (x) 
&= (\mathbf{B}^*_{\phi'_u} h'_{u}) (x) -
(\mathbf{B}^a_{\phi} h'_{u}) (x) \nonumber \\
& \le  \frac{2}{3} \varepsilon_2 +  (\mathbf{B}^*_{\phi(\set{C}_u)} h^*_{\phi(\set{C}_u)}) (x)
- (\mathbf{B}^a_{\phi} h^*_{\phi(\set{C}_u)}) (x) \nonumber  \\
& \le  \frac{2}{3} \varepsilon_2 - \varepsilon_2 = - \frac{1}{3} \varepsilon_2 \le \zeta_u 
\label{eq:delta-calc-with-errror}
\end{align}
where 
the second equality is from the construction of $\psi_u$
and the fact that $\set{O}(x;\phi'_u) \cap \set{A}^+ (x; \phi, \set{C}_u) = \emptyset$, i.e., $a \notin \set{O}(x;\phi'_u)$; and the first and second inequalities are from \eqref{eq:on-D-connect},
the definition of $\varepsilon_2$.
We have obtained that $\psi_u \in \Phi$ and $\delta^*(x,a;\psi_u, \set{C}_u, \zeta_u) = 0$ for some $a \notin \set{O}(x;\phi'_u)$). 
Therefore, $\psi_u \in \Delta_\Phi (\phi_u; \set{C}_u, \zeta_u)$.
Recalling the entering condition of the exploitation phase,
we establish the following relation:
\begin{align*} 
\set{D}_t(\zeta) \cap \set{M}'_t &\subseteq
\left\{
 \sum_{x \in \set{S}}\sum_{a \in \set{A}} 
{N_u (x,a)} \kl_{\phi_u \mid \psi_u}  (x,a)
 \ge \gamma_u
~\exists u \in [i^t_1, t]
  \right\}  \\
 &\subseteq
\left\{
 \sum_{x \in \set{S}}\sum_{a \in \set{A}} 
{N_u (x,a)}  \kl_{\phi_u \mid \phi}  (x,a)
 \ge \gamma_u
~\exists u \in [i^t_1, t]
  \right\} 
\end{align*}
where the last inclusion follows from the construction of $\psi_u$, i.e.,
\begin{align*}
 \sum_{x \in \set{S}}\sum_{a \in \set{A}} 
{N_u (x,a)}  \kl_{\phi_u \mid \psi_u}  (x,a)
& =  \sum_{x \in \set{S}}\sum_{a \notin \set{O}(x; \phi'_u)} 
{N_u (x,a)}  \kl_{\phi_u \mid \psi_u}  (x,a) \\
 & =  \sum_{x \in \set{S}}\sum_{a \notin \set{O}(x; \phi'_u)} 
{N_u (x,a)}  \kl_{\phi_u \mid \phi}  (x,a) \\
  & \le  \sum_{x \in \set{S}}\sum_{a \in \set{A}} 
{N_u (x,a)} \kl_{\phi_u \mid \phi}  (x,a) .
\end{align*}

As a consequence, applying the following lemma, $\Pr^\pi_{\phi \mid x_1} \left[\set{D}_t(\zeta), \set{M}'_t \right]$ is bounded by $o(1/t)$.
\begin{lemma} \label{lem:concentr}
Consider any $\pi$ and $\phi$ with Bernoulli rewards. Then, for any $\gamma >0$
and $\rho \in (0, 1)$, as $T \to \infty$,
\begin{align} \label{eq:kl-concentration}
\sum_{t = \rho T}^T \Pr\left[\sum_{x\in \set{S}} \sum_{a\in \set{A}} N_t(x,a) \kl_{\phi_t \mid \phi}(x,a) \ge (1+\gamma) \log t  \right]  = o(1/T).
\end{align}
\end{lemma}
{\bf Proof of Lemma~\ref{lem:concentr}.} 
The proof is an application of Theorem~2 in \cite{magureanu2014}, which says that for $\gamma' > SA+1$
and sufficiently large $t > 0$,
\begin{align*}
\Pr\left[\sum_{x\in \set{S}} \sum_{a\in \set{A}} N_t(x,a) \kl_{\phi_t \mid \phi}(x,a) \ge \gamma' \right]  \le e^{-\gamma'} \left( \frac{ (\gamma')^2 \log t }{SA}\right)^{SA} e^{SA+1} .
\end{align*}
Hence, putting $(1+\gamma) \log t$ to $\gamma'$,
we obtain
\begin{align*} 
&\sum_{t = \rho T}^T \Pr\left[\sum_{x\in \set{S}} \sum_{a\in \set{A}} N_t(x,a) \kl_{\phi_t \mid \phi}(x,a) \ge (1+\gamma) \log t  \right] \\
&\le 
\sum_{t = \rho T}^T 
e^{-(1+\gamma) \log t} \left( \frac{ (1+\gamma)^2 (\log t)^3 }{SA}\right)^{SA} e^{SA+1} \\
&\le 
\sum_{t = \rho T}^T 
e^{SA+1} \left( \frac{ (1+\gamma)^2}{SA}\right)^{SA} 
\frac{(\log t)^{3SA}}{t^{1+\gamma}} . 
\end{align*}
Using that ${(\log t)^{3SA}}/{t^{1+\gamma}}  = O({1}/{t^{1+\gamma/2}})$ for $t \ge \rho T$
and $\int_{\rho T}^T 1/t^{1+\gamma/2} dt \le 1/ (\rho T)^{1+\gamma/2} = o(1/T)$,
we conclude the proof of Lemma~\ref{lem:concentr}.
\ep

It remains to bound
$\Pr^\pi_{\phi \mid x_1} \left[\set{D}_t(\zeta), \neg\set{M}_t, \neg\set{M}'_t  \right]$.
From Lemma~\ref{lem:D-implies}, on the event $\set{D}_t(\zeta)$, 
it is true that $g^*_{u}$ is non-decreasing
in $u \in [i^t_1, t]$, i.e., $g^*_{i^t_k} \le g^*_{i^t_{k+1}}$.
Hence, 
\begin{align*}
\set{D}_t(\zeta) \cap \neg\set{M}_t ~\subseteq~ \set{D}_t(\zeta) \cap \left(\cup_{k = 1}^K \set{M}^k_t \right)
\end{align*}
where 
\begin{align*}
\set{M}^k_t  := \{ g^*_{i^t_{k}} = g^*_{i^t_{k+1}} < g^*_\phi \}.
\end{align*}
Then, it suffices to show that for any $k \in [1,K]$,
$\Pr^\pi_{\phi \mid x_1} \left[\set{D}_t(\zeta), \set{M}^k_t, \neg\set{M}'_t  \right] = o(1/t)$.

For $k \in [1, K]$, assume that the events $\set{D}_t(\zeta), \set{M}^k_t$ and $\neg\set{M}'_t$ occur. Fix $x \in \set{S}$ such that $\set{A}^+ (x; \phi, \set{C}_{i^t_k}) \neq \emptyset$. Since $g^*_{i^t_k} <  g^*_{\phi}$,  such a $x \in \set{S}$ must exist.
In addition, using the second part of Lemma~\ref{lem:D-implies}
and recalling \eqref{eq:D-implies-assum} with the fact that the ergodic MDPs $\phi(\set{C}_u), 
\phi(\set{C}_{u+1})$ have unique bias functions,
it follows that $g^*_{u} = g^*_{u+1}$ and $h^*_{u} = h^*_{u+1}~\forall u \in \set{I}^t_k$.
Therefore, $\set{A}^+ (x; \phi, \set{C}_{i^t_k}) = \set{A}^+ (x; \phi, \set{C}_{u}) \neq \emptyset ~\forall u \in [i^t_k, i^t_{k+1}]$. 
Recalling $N^t_k (x) := N_{i^t_{k+1}} (x) -  N_{i^t_{k}} (x) $
and $N^t_k (x, a) := N_{i^t_{k+1}} (x, a) -  N_{i^t_{k}} (x, a)$,
this implies that 
the algorithm never enters the exploitation phase when $X_u = x$, i.e.,
on the event $\{\set{D}_t(\zeta), \set{M}^k_t, \neg\set{M}'_t\}$,
\begin{align} 
N_{k+1}^t(x,a) &= \sum_{u \in \set{I}^t_k} \mathbbm{1} [(X_u, A_u) = (x,a), \neg\set{E}^\xpt_u] \nonumber \\
&= \sum_{u \in \set{I}^t_k} \mathbbm{1} [(X_u, A_u) = (x,a), \set{E}^\mnt_u \cup \set{E}^\est_u]
+\sum_{u \in \set{I}^t_k} \mathbbm{1} [(X_u, A_u) = (x,a), \set{E}^\xpr_u] \nonumber \\
&\le O(\log t)
+\sum_{u \in \set{I}^t_k} \mathbbm{1} [(X_u, A_u) = (x,a), \set{E}^\xpr_u] \label{eq:M-decomp-xpr}
\end{align}
where the last inequality is obtained since by Lemma~\ref{lem:monotone-bound}, the number of times the algorithm enters the monotonization phase is $O(\log t)$ (c.f., \eqref{eq:mnt-bound}),
and since by design, the algorithm limits the number of times we enter the estimation phase 
to $O(\log t/\log\log t)$.

Hence, it is enough to show that for $(x,a) \in \set{S} \times \set{A}$ such that 
$a \notin \set{A}^+(x ; \phi, \set{C}_{i^t_k}) \neq  \emptyset$,
\begin{align}
\Pr^\pi_{\phi \mid x_1} \left[\set{D}_t(\zeta), \set{M}^k_t, \neg\set{M}'_t,
 \set{L}^{k}_t (x,a) \right] = o(1/t) \label{eq:bound-3-final-final}
\end{align}
where we define
 $\set{L}^{k}_t (x,a) := \{ N^{t, \xpr}_k (x,a) \ge \frac{\rho \beta t}{2A} \}$
 with $N^{t, \xpr}_k (x,a) := \sum_{u \in \set{I}^t_k} \mathbbm{1} [(X_u, A_u) = (x,a),  \set{E}^\xpr_u]$.
Indeed, for $x \in \set{S}$ such that $\set{A}^+(x ; \phi, \set{C}_{i^t_k}) \neq  \emptyset$,
 if the event $\{\set{D}_t(\zeta), \set{M}^k_t, \neg\set{M}'_t,
\set{L}^{t}_k (x,a) ~\forall a \notin \set{A}^+(x ; \phi, \set{C}_{i^t_k}) \}$ occurs,
then 
\begin{align*}
\sum_{a \in \set{A}^+(x ; \phi, \set{C}_{i^t_k})} N^t_{k+1}(x,a) 
&=  N^t_{k+1}(x)  - 
\sum_{a \notin \set{A}^+(x ; \phi, \set{C}_{i^t_k})} N^t_{k+1}(x,a)  \\
& \ge \rho \beta t -  \sum_{a \notin \set{A}^+(x ; \phi, \set{C}_{i^t_k})} N^t_{k+1}(x,a)  \\
& \ge \rho \beta t - \sum_{a \notin \set{A}^+(x ; \phi, \set{C}_{i^t_k})} N^{t, \xpr}_{k}(x,a)  - O(\log t) \\
& \ge \rho \beta t - \frac{\rho \beta t}{2}  - O(\log t)
\end{align*}
where for the second inequality, we use \eqref{eq:M-decomp-xpr}.
This implies that 
for sufficiently large $t$, 
there exists 
$a \in \set{A}^+(x ; \phi, \set{C}_{i^t_k})$ such that $ N^t_{k+1}(x,a) \ge \frac{1}{3}\rho \beta t \ge \log^2 t \ge \log^2 {N_{i^t_{k+1}}(x)}$, i.e., $a \in \set{C}_{i^t_{k+1}}$,
and thus $g^*_{i^t_{k+1}}  > g^*_{i^t_{k}}$ which contradicts to the occurrence of the event $\set{M}^k_t$. Therefore, for sufficiently large $t > 0$
and $x \in \set{S}$ such that 
$\set{A}^+(x ; \phi, \set{C}_{i^t_k}) \neq  \emptyset$,
\begin{align*}
\Pr^\pi_{\phi \mid x_1} \left[\set{D}_t(\zeta), \set{M}^k_t, \neg\set{M}'_t  \right]
\le \sum_{a \notin \set{A}^+(x ; \phi, \set{C}_{i^t_k})}
\Pr^\pi_{\phi \mid x_1} \left[\set{D}_t(\zeta), \set{M}^k_t, \neg\set{M}'_t,
\set{L}^{k}_t (x,a)   \right] . 
\end{align*}

It remains to prove \eqref{eq:bound-3-final-final}.
Fix $(x,a) \in \set{S} \times \set{A}$ such that $ a \notin \set{A}^+(x ; \phi, \set{C}_{i^t_k}) \neq  \emptyset$,
 Assume that
the events $\set{D}_t(\zeta), \set{M}^k_t, \neg\set{M}'_t$ occur and 
$N^{t, \xpr}_k (x,a) \ge \frac{\rho \beta t}{2A}$.
Let 
\begin{align*}
t_3 := \min \left\{u \in \set{I}^t_k : 
\sum_{v = i^t_k}^{u} \mathbbm{1} [(X_v, A_v) = (x,a),  \set{E}^\xpr_v]
 \ge \frac{\rho \beta t}{4A} \right\}.
\end{align*}
From the assumption, $t_3 \in \set{I}^t_k$.
Then, using a similar argument as that used to derive \eqref{eq:bound-3-second-sum} and using Lemma~\ref{lem:concen-exp}, 
we can guarantee $\| \phi_u(x,a) - \phi(x,a) \| \le \zeta$ 
for all $u  > t_3$ with probability $1-o({1}/{t})$
as $N_{u}(x,a) \ge \frac{\rho \beta t}{4A} = \Omega(t) $
, i.e.,
\begin{align}
\Pr^\pi_{\phi \mid x_1} \left[\set{D}_t(\zeta), \set{M}^k_t, \neg\set{M}'_t,
\set{L}^{k}_t (x,a), 
\| \phi_u(x,a) - \phi(x,a) \| > \zeta
~\exists~u \in [t_3, i^t_{k+1}]  \right] = o(1/t).  \label{eq:bound-3-final-final-concentr}
\end{align}
Further, assume that $\| \phi_u(x,a) - \phi(x,a) \| \le \zeta, \ \forall u \in [t_3, i^t_{k+1}]$.
Then, similarly as in \eqref{eq:delta-calc-with-errror}, we can deduce that $\delta^*(x,a;\phi_u, \set{C}_u) > \zeta_u $ from the assumption of the correctness of the estimated bias function and \eqref{eq:on-D-connect}, and thus 
\begin{align*}
\delta^*(x,a;\phi_u, \set{C}_u, \zeta_u) > \zeta_u .
\end{align*}
Hence, in the exploration phase, 
when $\set{F}_u = \emptyset$, $\eta_u(x,a) = 0$ due to the design of the algorithm,
while when $\set{F}_u \neq \emptyset$, $\eta_u(x,a) \le 2SA \left(\frac{S+1}{\zeta_u} \right)^2$ due to Lemma~\ref{lem:sol-bound}.
Therefore, 
recalling the definition of $\gamma'_t$ in \eqref{eq:def-gamma_t-prime}, 
for any $u \in [t_3, i^t_{k+1}]$, on the event $\{\set{E}^\xpr_u, X_u = x\}$,
\begin{align*}
\eta_u(x,a) \gamma_u \le 2SA \left(\frac{S+1}{\zeta_u} \right)^2 \gamma_u \le 
\gamma'_t= O(\log^2 t)
\end{align*} 
which implies that for sufficiently large $t > 0$ such that
 $\gamma'_t = O(\log^2 t) < \frac{\rho \beta t }{4A} \le N_u(x,a) = \Omega(t)$, $\mathbbm{1} [\set{E}^\xpr_u, (X_u, A_u) = (x,a)] = 0$
for all  $u \in [t_3, i^t_{k+1}]$ due to the design of the exploration phase.
Hence, it follows that
\begin{align*}
\Pr^\pi_{\phi \mid x_1} \left[\set{D}_t(\zeta), \set{M}^k_t, \neg\set{M}'_t,
\set{L}^{k}_t (x,a), 
\| \phi_u(x,a) - \phi(x,a) \| \le \zeta
~\forall~u \in [t_3, i^t_{k+1}]  \right] = 0. 
\end{align*}
Combining the above with \eqref{eq:bound-3-final-final-concentr}, 
we have completed the proof of \eqref{eq:bound-3-final-final} 
and thus the proof of Lemma~\ref{sec:pf-difficult-bound-3}.

\ep


%

\end{document}